\pdfoutput=1

\documentclass[11pt]{article}

\usepackage[final]{emnlp2021}

\usepackage{times}
\usepackage{latexsym}

\usepackage{amsmath, amssymb, amsthm, amscd, amsfonts}
\usepackage{IEEEtrantools}
\usepackage{graphicx}
\usepackage{caption}
\usepackage{subcaption}
\usepackage{booktabs, tabularx, ragged2e}

\usepackage[T1]{fontenc}

\usepackage[utf8]{inputenc}

\usepackage{microtype}

\theoremstyle{definition}

\theoremstyle{remark}

\newcommand{\RN}[1]{%
  \textup{\uppercase\expandafter{\romannumeral#1}}%
}

\DeclareMathOperator*{\argmin}{arg\,min}

\title{Assessing the Reliability of Word Embedding Gender Bias Measures}

\author{
  Yupei Du, Qixiang Fang and Dong Nguyen\\
  Utrecht University\\
  Utrecht, the Netherlands \\
  \texttt{\{y.du,q.fang,d.p.nguyen\}@uu.nl} \\
  }

\begin{document}
\maketitle
\begin{abstract}
  Various measures have been proposed to quantify 
  human-like social biases in word embeddings. 
  However, 
  bias scores based on these measures can suffer from measurement error. 
  One indication of measurement quality is  \emph{reliability}, 
  concerning the extent to which a measure produces consistent results. 
  In this paper, 
  we assess three types of reliability 
  of word embedding gender bias measures, 
  namely test-retest reliability, inter-rater consistency 
  and internal consistency. 
  Specifically, we investigate
  the consistency of bias scores across 
  different choices of random seeds, scoring rules and words.
  Furthermore, we 
  analyse 
  the effects of various 
  factors 
  on these measures' reliability scores. 
  Our findings 
  inform better design of word embedding gender bias measures. 
  Moreover, 
  we urge researchers to be more critical about 
  the application of such measures.\footnote{
    Our code is available at \url{https://github.com/nlpsoc/reliability_bias}.} 

\end{abstract}

\section{Introduction}\label{sec:intro}

Despite their success in various applications, 
word embeddings have been shown to exhibit a range of human-like social biases. 
For example, 
\citet{caliskan2017semantics} find that both 
GloVe \citep{pennington-etal-2014-glove} and skip-gram \citep{mikolov-2013-word2vec} 
embeddings associate pleasant terms (e.g. \emph{love} and \emph{peace}) more
with European-American names than with African-American names, and that they
 associate career words (e.g. \emph{profession} and \emph{business}) more
with male names than with female names.

Various measures have been proposed 
to quantify such biases in word embeddings
\citep{DBLP:conf/acl/EthayarajhDH19,zhou-etal-2019-examining,
    manzini-etal-2019-black}.
These measures 
allow us to assess biases in word embeddings and 
the performance of 
bias-mitigation methods
\citep{bolukbasi2016man,zhao-etal-2018-learning}.
They also enable 
us
to study social biases in a new way 
complementary to 
traditional qualitative and experimental approaches 
\citep{garg2018word,chaloner-maldonado-2019-measuring}.

\begin{figure}[t]
    \centering
        \includegraphics[width=0.48\textwidth]{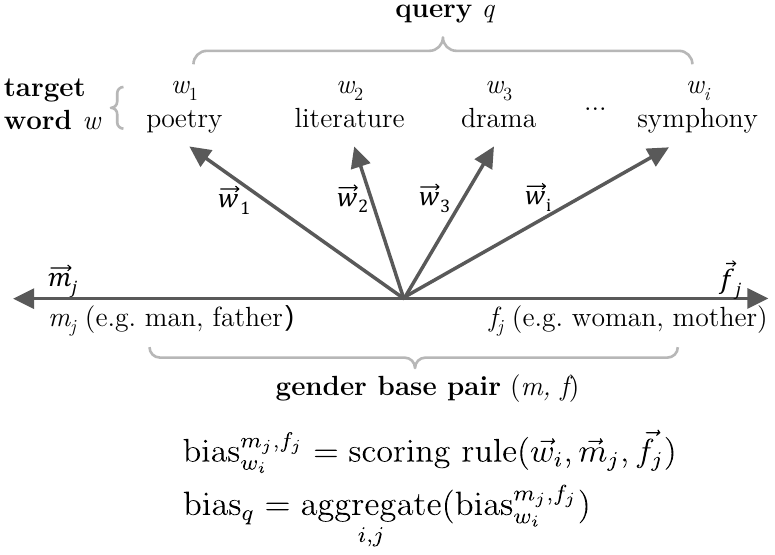}
    \caption{Measuring word embedding gender biases in the concept of \emph{arts}. 
        The arrows are hypothetical low-dimensional projections 
        of word embeddings.}
    \label{fig1}
\end{figure}

A key challenge in developing bias measures is that 
social biases are
abstract concepts that cannot be measured directly but have to be inferred from some observable data.
This renders the resulting bias scores more prone to measurement errors. 
Therefore, it is important to carefully assess 
these measures' measurement quality. 
\citet{abigail2021maf} also highlight similar measurement issues 
in the context of fairness research.

In this paper, we focus on one aspect of measurement quality: \textbf{reliability}. 
It concerns the extent to which 
a measure produces consistent results.
In particular, we investigate the reliability of word embedding 
\textbf{gender bias} measures.

Figure \ref{fig1} illustrates 
how gender biases in word embeddings are typically measured, 
with measuring gender biases in the concept of \emph{arts} as an example. 
To calculate the gender bias score of a \textbf{target word} $w$ 
(e.g. \emph{poetry}, a word that relates to the concept of interest), 
we need to specify a \textbf{gender base pair} $(m, f)$
(a gendered word pair, 
e.g. \emph{father/mother})
and \textbf{a scoring rule}. 
A scoring rule is a function that takes the embedding vectors of 
$w$, $m$ and $f$ as input 
and returns a bias score $bias_{w}^{m,f}$ as output. 
In practice, often 
an \textbf{ensemble of gender base pairs} are used. 
In this case, we aggregate (e.g. average)
all the bias scores w.r.t. different gender base pairs 
to obtain an overall bias score for a target word. 
Furthermore, 
multiple conceptually related target words 
(e.g. \emph{poetry},  
\emph{drama}) 
may be specified to form a \textbf{query} $q$ (e.g. \emph{arts}). 
By aggregating individual bias scores of these target words, 
we can compute an overall bias score for a concept.

Clearly, the choice of target words, 
gender base pairs and scoring rules 
may influence the resulting bias scores. 
Reliable measurements of word embedding gender biases 
thus require 
target words, gender base pairs and scoring rules that 
produce consistent results. 
In this work, 
by drawing from measurement theory, 
we propose a comprehensive approach (\S \ref{sec:reliability})
to evaluate three types of reliability 
for these different components:

\begin{itemize}
    \setlength\itemsep{0.1em}
    \item First, 
        we assess 
        the consistency of bias scores associated with different 
        target words, gender base pairs and scoring rules,  
        over different random seeds 
        used in word embedding training 
        (i.e. \textbf{test-retest reliability}; \S \ref{subsec:test_retest}). 
    \item Second, 
        we assess  
        the consistency of bias scores associated with 
        different target words and gender base pairs, 
        across different scoring rules 
        (i.e. \textbf{inter-rater consistency}; \S \ref{subsec:inter_rater}). 
    \item Third, we assess 
        the consistency of bias scores 
        over 1) different target words within a query and 
        2) different gender base pairs 
        (i.e. \textbf{internal consistency}; \S \ref{subsec:internal}). 
\end{itemize}

Furthermore, we use multilevel regression  
to model the effects of  
various factors (e.g. word properties, embedding algorithms, training corpora) 
on the reliability scores of target words (\S \ref{subsec:factors}).

Our experiments show that word embedding gender
bias scores are mostly consistent across different random seeds (i.e. high test-retest reliability) and 
across target words within the same query (i.e. high internal consistency). 
However, different scoring rules generally fail to agree with one another (i.e. low inter-rater consistency).
Moreover, word embedding algorithms have a large influence on the reliability of bias scores.

\paragraph{Contributions} 
First, we connect measurement theory to word embedding bias measures. 
Specifically, we propose a reliability evaluation framework for word embedding (gender) bias measures. 
Second, we provide a comprehensive assessment of the reliability of word embedding gender bias measures. 
Based on our findings, we urge researchers to be more critical about applying such measures.

\section{Related Work}

Measuring gender biases in word embeddings has been receiving a growing amount of research interest in NLP. 
Various gender bias measures have been proposed.
They are based on different techniques, 
such as linear gender subspace identification \citep{
    bolukbasi2016man,vargas-cotterell-2020-exploring,
    manzini-etal-2019-black}, 
psychological tests \citep{DBLP:conf/acl/EthayarajhDH19,
    caliskan2017semantics}, 
inference from nearest neighbours \citep{DBLP:conf/naacl/GonenG19} 
and regression \citep{sweeney2019transparent,wefe2020}. 

However, recent studies have raised concerns 
over the reliability of such measures. 
\citet{zhang2020robustness} show that gender bias scores easily
vary in their direction and magnitude
when different forms (e.g. capitalisation) of target words or
different gender base pairs are used. 
Similarly, \citet{antoniak-mimno-2021-bad} look into 178 different gender base pairs from previous works
and find that the choice of gender base pairs can greatly impact bias measurements.
They therefore urge future work to examine and document the choices of gender base pairs. 
Moreover, 
\newcite{d2020underspecification} 
find that underspecification of models can lead to
unstable contextualised word embedding bias scores. 
These findings call for a more systematic evaluation of the
reliability of word embedding gender bias measures, 
which is the goal of our study.

Such measures' lack of reliability may partly stem from 
the fact that word embeddings themselves are often unstable, sensitive to
choices of, for instance, 
word embedding algorithms \citep{burdick2018factors,antoniak2018evaluating,
    hellrich-etal-2019-influence}, 
hyper-parameters \citep{levy2015improving,mimno2017strange,
    hellrich-etal-2019-influence} 
and even random seeds \citep{burdick2018factors,
    DBLP:conf/coling/HellrichH16,bloem-etal-2019-evaluating}
during word embedding training.

Various word (embedding) attributes have been found to 
contribute to the instability of word embeddings, 
including 
part-of-speech tags \citep{burdick2018factors,
    pierrejean-tanguy-2018-predicting}, 
word frequency \citep{DBLP:conf/coling/HellrichH16,
    pierrejean-tanguy-2018-predicting} 
and 
word ambiguity \citep{burdick2018factors,DBLP:conf/coling/HellrichH16}.

\section{Preliminaries}
In this section, we first review three popular scoring rules used for measuring  
word embedding gender biases (\S \ref{sec:bias_metrics}).
Then, we introduce the conceptual framework of reliability 
and motivate its use in word embedding gender bias measurements (\S \ref{sec:statistical_reliability_ss}).

\subsection{Scoring Rules}\label{sec:bias_metrics}

Following \citet{zhang2020robustness}, 
we focus on three popular scoring rules:
DB/WA, RIPA and NBM.\footnote{
This terminology is also adopted from \citet{zhang2020robustness}.}

\paragraph{DB/WA}
DB/WA (Direct Bias / Word Association) is 
one of the most commonly used scoring rules in previous work 
\citep{bolukbasi2016man,caliskan2017semantics}.
Given a gender base pair $(m, f)$, 
the DB/WA score of a target word $w$ is 
\begin{IEEEeqnarray}{rCl}
    \text{DB/WA}^{(m, f)}_{w} = 
    \cos{(\vec w, \vec m)} - \cos{(\vec w, \vec f)}, \notag
\end{IEEEeqnarray}
where $\vec{*}$ is the corresponding word vector of $*$, 
and $\cos(x, y)$ refers to the cosine similarity of $x$ and $y$. 

\paragraph{RIPA}
Another scoring rule based on vector similarity 
is Relational Inner Product Association 
\citep[RIPA;][]{DBLP:conf/acl/EthayarajhDH19}. 
The main difference between DB/WA and RIPA is that 
RIPA performs normalisation at the 
gender base pair level instead of at the word level. 
Formally, 
\begin{IEEEeqnarray}{rCl}
    \text{RIPA}^{(m, f)}_{w} = 
    \vec w \cdot \frac{\vec m - \vec f}
    {\|\vec m - \vec f\|}, \notag
\end{IEEEeqnarray}
where $\|*\|$ refers to the L2 norm of $*$. 

\paragraph{NBM}
Unlike DB/WA and RIPA, which are based on vector similarities, 
NBM (Neighbourhood Bias Metric) is based on 
a word's $k$ nearest neighbours \citep{DBLP:conf/naacl/GonenG19}. 
Specifically, 
\begin{IEEEeqnarray}{rCl}
\small
    \text{NBM}^{(m, f)}_{w} = 
    \frac{|masculine(w)| - |feminine (w)|}{k}\notag, 
\end{IEEEeqnarray}
where $|masculine(w)|$ and $|feminine(w)|$ are 
the number of words in $w$'s $k$ nearest neighbours  
biased towards the respective gender based on their DB/WA scores.
Following \citet{zhang2020robustness} and \citet{DBLP:conf/naacl/GonenG19}, 
we use $k=100$. 

\subsection{Reliability}
\label{sec:statistical_reliability_ss}
In measurement theory, 
reliability is the extent to which a measure 
produces consistent results over a variety of measurement conditions,  
in which basically the same results should be obtained \citep{drost_validity_2011}.
In this work, we focus on three important types of reliability:
test-retest reliability, inter-rater consistency and internal consistency.

\textbf{Test-retest reliability} concerns the consistency of measurements
across different measurement occasions 
(assuming no substantial change in the true value, \citealp{weir_quantifying_2005}).
For example, 
if gender bias scores vary substantially 
across different measurement occasions 
(e.g. different random seeds during embedding training; 
    different random data samples), 
they should be considered to have low 
test-retest reliability.
In this case, derived conclusions 
from these scores 
are likely to be untrustworthy.

\textbf{Inter-rater consistency} is 
the degree to which different raters produce consistent measurements \citep{shrout_intraclass_1979}.
For example, 
consider scoring rules as the raters of word embedding bias scores. 
In this case, if different scoring rules measure gender biases in a similar way, 
they should produce bias scores 
that tend to agree with one another in both signs and normalised magnitude.

\textbf{Internal consistency} is defined as the agreement 
among multiple components that make up a measure of a single construct \citep{cronbach1951coef}.
In the example from Figure~\ref{fig1}, 
we specify a query consisting of 
various \emph{arts}-related target words to measure gender biases 
of the concept \emph{arts}.
We then compute individual bias scores 
for all target words before aggregating them to obtain an overall bias score.
If the bias scores of target words are distinct from one another 
(i.e. low or negative correlation), 
the query has low internal consistency. 
In this case, 
one should question whether these target words measure the
\emph{arts} concept in comparable ways.

\begin{table*}[t]
    \small
    \begin{tabularx}{\textwidth}{XXXXXXX}
    \toprule
              & \multicolumn{2}{c}{\textbf{Test-retest reliability}} & \multicolumn{2}{c}{\textbf{Inter-rater consistency}} & \multicolumn{2}{c}{\textbf{Internal consistency}}\\ \midrule

    Component & Target words & Gender base pairs & Target words & Gender base pairs  & Gender base pair ensemble & Queries \\ \hline

    Source of variations & \multicolumn{2}{c}{Different random seeds} & \multicolumn{2}{c}{Different scoring rules} & Individual gender base pairs & Individual target words \\
    \bottomrule
    \end{tabularx}
    \caption{Operational definitions for different types of reliability.}
    \label{tab:reliability}
    \end{table*}

\section{Estimating Reliability of Word Embedding Gender Bias Measures}
\label{sec:reliability}
In this section, 
we propose an evaluation framework to assess
the reliability of word embedding gender bias measures. 
Respectively, we present our operational definitions 
for test-retest reliability (\S \ref{sec:test_retest_emb}),
inter-rater consistency (\S \ref{sec:inter_rater_emb}) and
internal consistency (\S \ref{sec:internal_consistency_emb}). 
See Table \ref{tab:reliability} for an overview.

\paragraph{Notation} 
Suppose we have $s$ scoring rules, 
$g$ gender base pairs and $t$ target words. 
We train $k$ word embedding models 
with the same hyper-parameters but $k$ different random seeds. 
For each scoring rule, 
we calculate the bias score of each target word 
w.r.t each gender base pair, 
on each word embedding model. 
As a result, 
we get a four-dimensional bias scores matrix $B$ 
 of $\mathbb{R}^{s \times g \times t \times k}$. 
For calculating inter-rater consistency and internal consistency, 
we average the gender bias scores derived from 
the $k$ word embedding models. 
Averaging these embedding models 
can partial out the influence of random seeds, 
and therefore lead to more accurate estimation of 
other types of reliability. 
In this way, we get another bias matrix 
$B^{\prime}$ of $\mathbb{R}^{s \times g \times t}$.

\subsection{Test-retest Reliability}
\label{sec:test_retest_emb}

We measure test-retest reliability as the consistency of bias scores associated with each target word, gender base pair and scoring rule across different random seeds\footnote{
    One may argue that we can use 
    pre-trained word embeddings. 
    However, deterministic alternatives 
    are not free of reliability issues. 
    Instead, they are of ``fixed randomness''
    \citep{DBLP:conf/coling/HellrichH16}. 
}. 
Specifically, 
we focus on two questions: 
First, are the bias scores of 
a single target word (averaged across gender base pairs) consistent 
across random seeds (i.e. \textbf{test-retest reliability of target words})? 
Second, are the average bias scores of all target words w.r.t one gender base pair consistent
across random seeds (i.e. \textbf{test-retest reliability of gender base pairs})?
We explore both questions for each scoring rule separately. 

To calculate test-retest reliability for each target word, 
we use a bias score matrix of size $\mathbb{R}^{g \times k}$ 
by slicing $B$. 
We then use intra-class correlation (ICC) 
to estimate test-retest reliability. 

ICC is a popular family of estimators for  
both test-retest reliability and inter-rater consistency. There are different forms of ICC estimators, 
each of which can involve distinct assumptions and 
can therefore lead to very different interpretations \cite{koo2016guideline}.
\citet{shrout_intraclass_1979}
define 6 forms of ICC and present them as ``ICC'' with 2 additional numbers in parentheses (e.g., ICC(2,1) and ICC(3,1)). 
The first number refers to the \textbf{model} and can take on three possible values (1, 2 or 3).
The second number refers to the \textbf{intended use} of raters/measurements in an application and can take on two values (1 or k). 
See Appendix \ref{app:ICC} for a detailed description of these value options.
We adopt ICC(2,1) as the estimator for test-retest reliability: 

\small
\begin{IEEEeqnarray}{rCl}\label{eq:icc_21}
\mathcal{I}^{(2, 1)} = 
\frac{ M S_{R}-M S_{E} }{M S_{R}+
(k-1) M S_{E}+\frac{k}{t}\left(M S_{C}-M S_{E}\right)},
\end{IEEEeqnarray}
\normalsize
where $M S_{E}$, $M S_{R}$, and $M S_{C}$ are 
the mean square of error, of rows and of columns, respectively\footnote{See our code for the exact implementation of the different reliability estimators}. 

Similarly, 
for the test-retest reliability of each gender base pair, 
we slice $B$ to get a bias score matrix of size $\mathbb{R}^{t \times k}$. 
We then calculate its ICC value using Equation \ref{eq:icc_21}
(by substituting $t$ with $g$). 

\subsection{Inter-rater Consistency}
\label{sec:inter_rater_emb}
We define inter-rater consistency as
the consistency of bias scores 
across different scoring rules. 
We again investigate two questions: 
First, are the bias scores of 
a single target word (averaged across all gender base pairs) consistent 
across scoring rules  
(i.e. \textbf{inter-rater consistency of a target word})? 
Second, are the average bias scores of all
target words w.r.t. a single gender base pair 
consistent
across scoring rules 
(i.e. \textbf{inter-rater consistency of a gender base pair})?

To calculate the inter-rater consistency of each target word, 
we use a bias score matrix of size $\mathbb{R}^{g \times s}$ 
by slicing and transposing $B^{\prime}$. 
Following \citet{koo2016guideline},
we adopt ICC(3, 1) as the estimator:

\begin{IEEEeqnarray}{rCl}\label{eq:icc_31}
\mathcal{I}^{(3, 1)} = 
\frac{ M S_{R}- M S_{E}}{M S_{R}+(s-1)  M S_{E}}.
\end{IEEEeqnarray}

Similarly, for the second question, 
we can get a bias score matrix size of $\mathbb{R}^{t \times s}$ 
for each gender base pair and calculate the inter-rater consistency of 
the gender base pair via Equation \ref{eq:icc_31}. 

\subsection{Internal Consistency}
\label{sec:internal_consistency_emb}

We investigate the internal consistency of 
both queries and the ensemble of gender base pairs.
We thus focus on two questions: 
First, are the bias scores 
of different target words within a query consistent
(i.e. \textbf{internal consistency of a query})?
Second, are the average bias scores of all
target words consistent across different gender base pairs 
(i.e. \textbf{internal consistency of the ensemble of gender base pairs})? We examine both questions for each scoring rule separately.

To calculate the internal consistency of a query consisting of $t$ target words, 
we first slice and transpose $B^{\prime}$ to get 
a bias score matrix size of $\mathbb{R}^{g \times t}$. 
We then use Cronbach's alpha \cite{cronbach1951coef} 
as the estimator of internal consistency. 
Cronbach's alpha is the most common estimator 
for internal consistency, 
which assesses how closely related a set of test items are as a group 
(e.g. different target words of the same query). 
Specifically, 

\begin{IEEEeqnarray}{rCl}\label{eq:cronbach_alpha}
    \alpha=
    \frac{t}{t-1}\left(1-\frac{\sum_{i=1}^{t} 
    \sigma_{i}^{2}}{\sigma_{X}^{2}}\right), 
\end{IEEEeqnarray}
where 
$\sigma_{i}^{2}$ is the variance of the bias scores of 
target word $i$ in the query w.r.t. different gender base pairs. 
$\sigma_{X}^{2}$ is the sum of $\sum_{i=1}^{t} 
    \sigma_{i}^{2}$
and all covariances of bias scores between target words. 

We calculate the internal consistency of the ensemble of gender base pairs
in a similar way, 
by creating a bias score matrix size of $\mathbb{R}^{t \times g}$ 
and applying Equation \ref{eq:cronbach_alpha} 
(substituting $t$ with $g$). 

\section{Experiments}
\subsection{Experimental Setup}

\paragraph{Training Embeddings}
We select three corpora with different characteristics   
to train word embeddings. 
Two are from subReddits: r/AskScience ($\sim$ 158 million tokens) 
and r/AskHistorians 
($\sim$ 137 million tokens, also used by
\citealt{antoniak2018evaluating})\footnote{
    We use all posts and replies from 2014, 
    retrieved from \url{https://archive.org/download/2015_reddit_comments_corpus/reddit_data/2014/}.}. 
The third is the training set of WikiText-103 
($\sim$ 527 million tokens, \citealp{merity2016pointer}), 
consisting of high-quality Wikipedia articles.

We use two popular word embedding algorithms: 
Skip-Gram with Negative Sampling (SGNS; \citealt{mikolov-2013-word2vec}) 
and GloVe \citep{pennington-etal-2014-glove}. 
For both algorithms, we set the number of embedding dimensions to 300. 
For all other hyper-parameters, 
we use the default values of previous implementations.\footnote{
    For SGNS, we use Gensim 3.8.3 \citep{rehurek_lrec} 
    with a window size of 5, a minimum word count of 5 and 
    5 iterations.  
    For GloVe, we use the official implementation 
    \url{https://github.com/stanfordnlp/GloVe},
    with a window size of 15, a minimum word count of 5 and 
    15 iterations.  
    Because we do not fine-tune hyper-parameters, 
    our results do not necessarily indicate 
    which algorithm itself is of better reliability. 
    To investigate the potential impact of this decision, 
    we also include an explorative study on the influence of hyper-parameters 
    in Appendix \ref{app:hyperparameter}. 
}
For each corpus-algorithm pair, 
we train $k=32$ word embedding models using different random seeds. 

\paragraph{Gender Base Pairs}
We collect and lower-case all 23 gender base pairs from 
\citet{bolukbasi2016man} and \citet{garg2018word}.

\paragraph{Target Word Lists}
For the assessment of test-retest reliability and inter-rater consistency, 
we include three word lists used in previous word embedding bias studies: 
1) 320 occupation words from \citet{bolukbasi2016man} (OCC16),
2) 76 occupation words from \citet{garg2018word} (OCC18) and
3) 230 adjectives from \citet{garg2018word} (ADJ).
However, these three lists are very specific (i.e. only concerning occupation words and adjectives) and thus unlikely applicable to other (future) research where different biases are of interest and different target words might be used (e.g. measuring gender biases of a whole corpus).
Therefore, 
we also consider two additional, larger target word lists: 
4) the top 10,000 most frequent words of  
Google's trillion word corpus
(Google10K)\footnote{\url{https://github.com/first20hours/google-10000-english}.} and
5) the full vocabulary of each corpus (Full).

\paragraph{Queries}
For the assessment of internal consistency, 
we examine six gender bias related queries from 
\citet{caliskan2017semantics}: \emph{math}, \emph{arts}, \emph{arts\_2}, 
\emph{career}, \emph{science}, and \emph{family}, 
each consisting of eight target words. Note that target word lists are different from queries. 
    The former does not necessarily consist of conceptually related words.

\subsection{Results: Test-retest Reliability}\label{subsec:test_retest}

\begin{figure}[t]
    \centering
    \begin{subfigure}{0.48\textwidth}
        \centering
        \includegraphics[width=\textwidth]{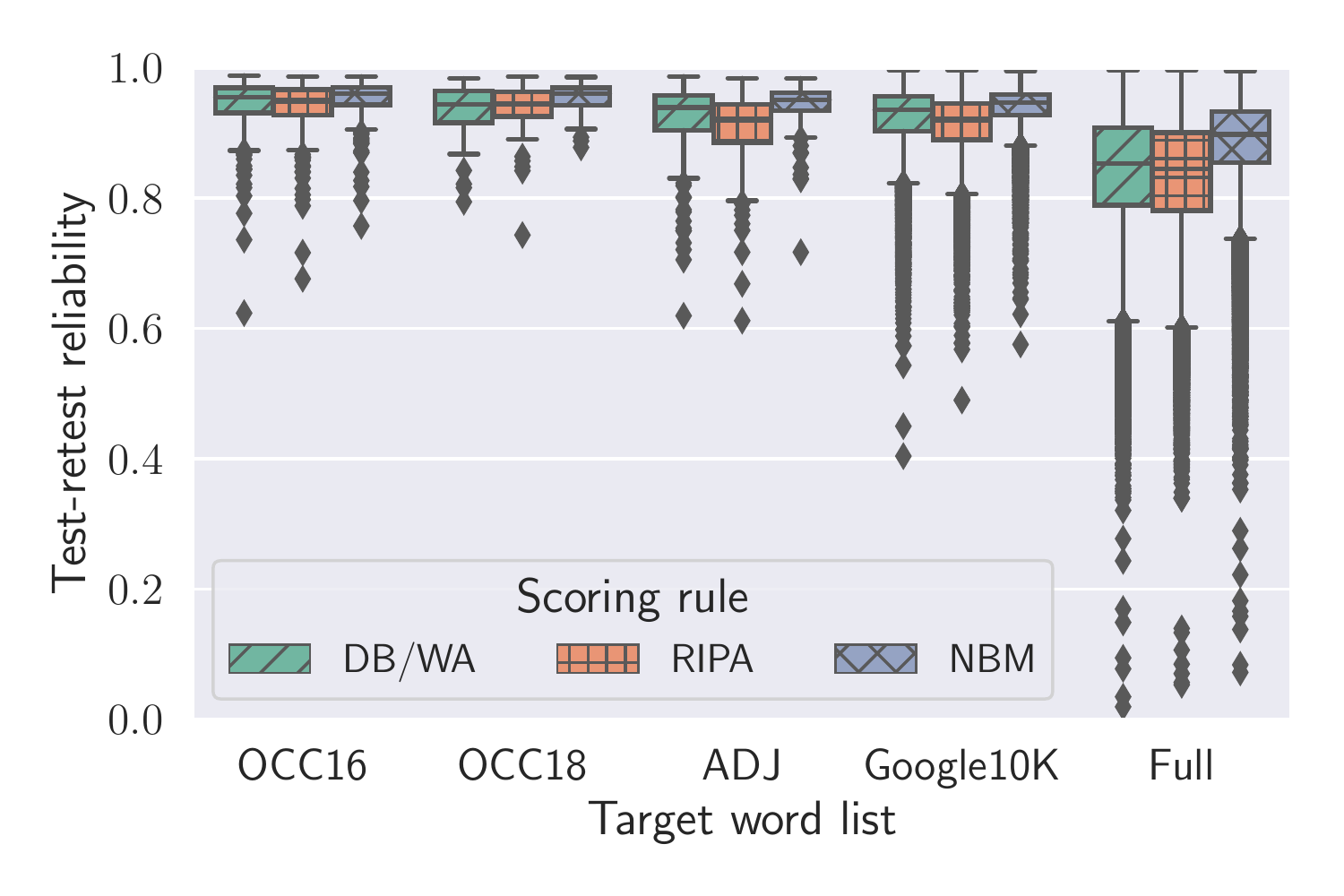}
        \caption{Target words}
        \label{subfig:retest_target}
    \end{subfigure}
    \vfill
    \begin{subfigure}{0.48\textwidth}
        \centering
        \includegraphics[width=\textwidth]{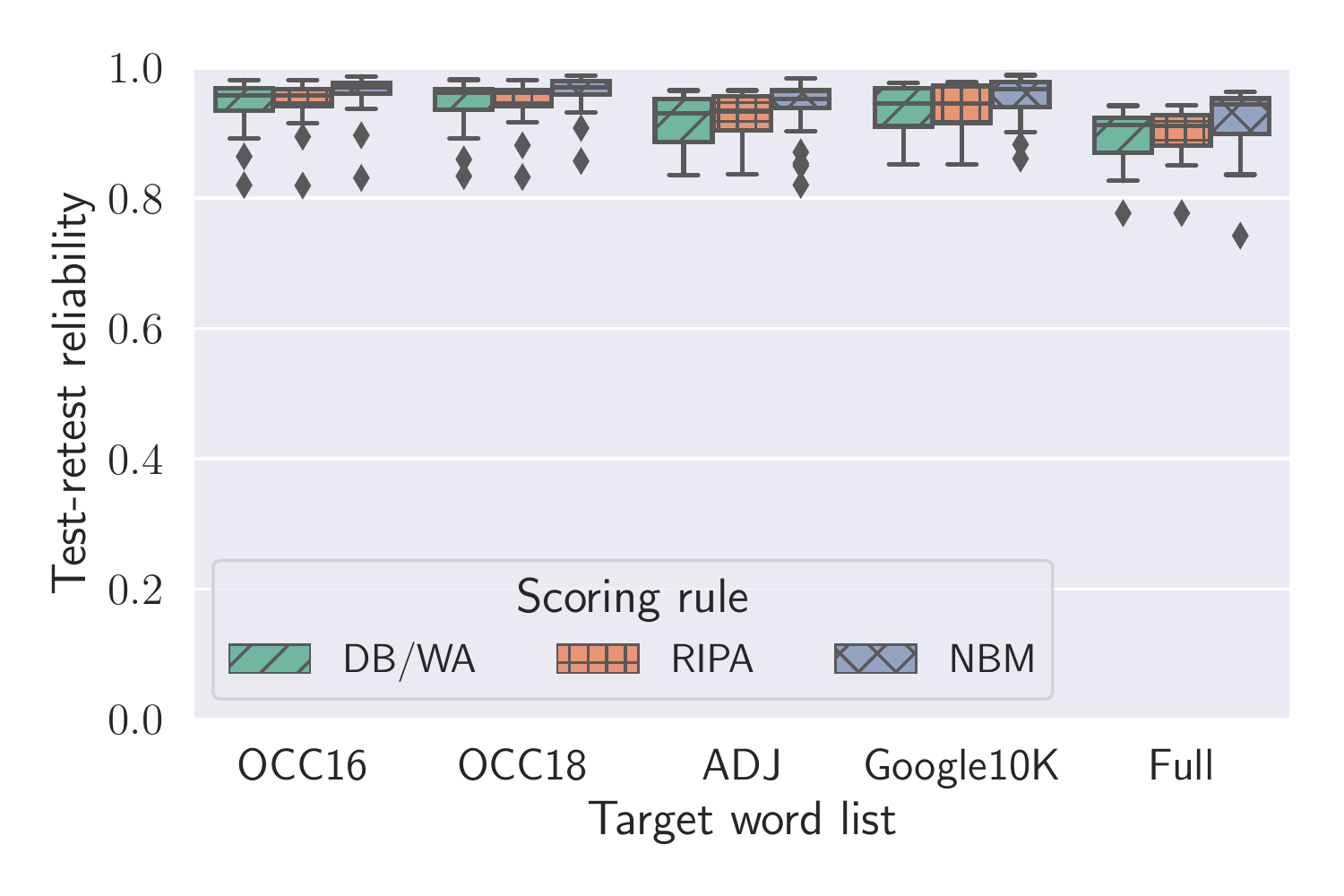}
        \caption{Gender base pairs}
        \label{subfig:retest_gbp}
    \end{subfigure}
    \caption{Distribution of the test-retest reliability scores
        across target word lists and scoring rules (based on SGNS and WikiText-103). Most target words and gender base pairs have acceptable test-retest reliability across the scoring rules. Nevertheless, some outliers with low test-retest reliability exist, especially for target word lists with infrequent words.}
    \label{fig:retest_box}
\end{figure}

Figure \ref{fig:retest_box} shows the distribution of  test-retest reliability scores of target words and gender base pairs across target word lists and scoring rules.
Here, word embeddings are trained with SGNS on WikiText-103. 
Similar results are found for other corpora and algorithms 
(see Appendix \ref{app:test_retest_reliability_figure}). 

First, we observe that the majority of target words and gender base pairs  
have acceptable test-retest reliability 
with ICC values greater than 0.6, 
regardless of the scoring rule used.\footnote{
    Generally, ICC values less than 0.5, between 0.5 and 0.75, 
    between 0.75 and 0.9, and greater than 0.9 
    are considered to have poor, moderate, good, and excellent 
    reliability, respectively \citep{koo2016guideline}.
    This also holds for ICC values of inter-rater consistency 
    (Section \ref{subsec:inter_rater}). 
}
Nevertheless, quite some target words and gender base pairs 
fall below the lower whiskers of the box-plots 
(indicating low test-retest reliability). 

Moreover, comparing with Google10K, 
which consists of frequent words, 
a higher proportion of words in the full vocabulary have 
very low test-retest reliability. 
For example, 0.01\% of the target words in Google10K have 
a test-retest reliability lower than 0.50
for word embeddings trained with SGNS on WikiText-103.
In contrast, for the full vocabulary this is 0.1\%, 
approximately 10 times that of Google10K. 
These results suggest that we should be careful when
making word lists that consist of infrequent words 
(e.g. when studying less common concepts). 
If we do need to use infrequent words, 
we should check their test-retest reliability 
before deriving further conclusions.

\subsection{Results: Inter-rater Consistency}\label{subsec:inter_rater}

\begin{figure}[t]
    \centering
    \begin{subfigure}{0.48\textwidth}
        \centering
        \includegraphics[width=\textwidth]{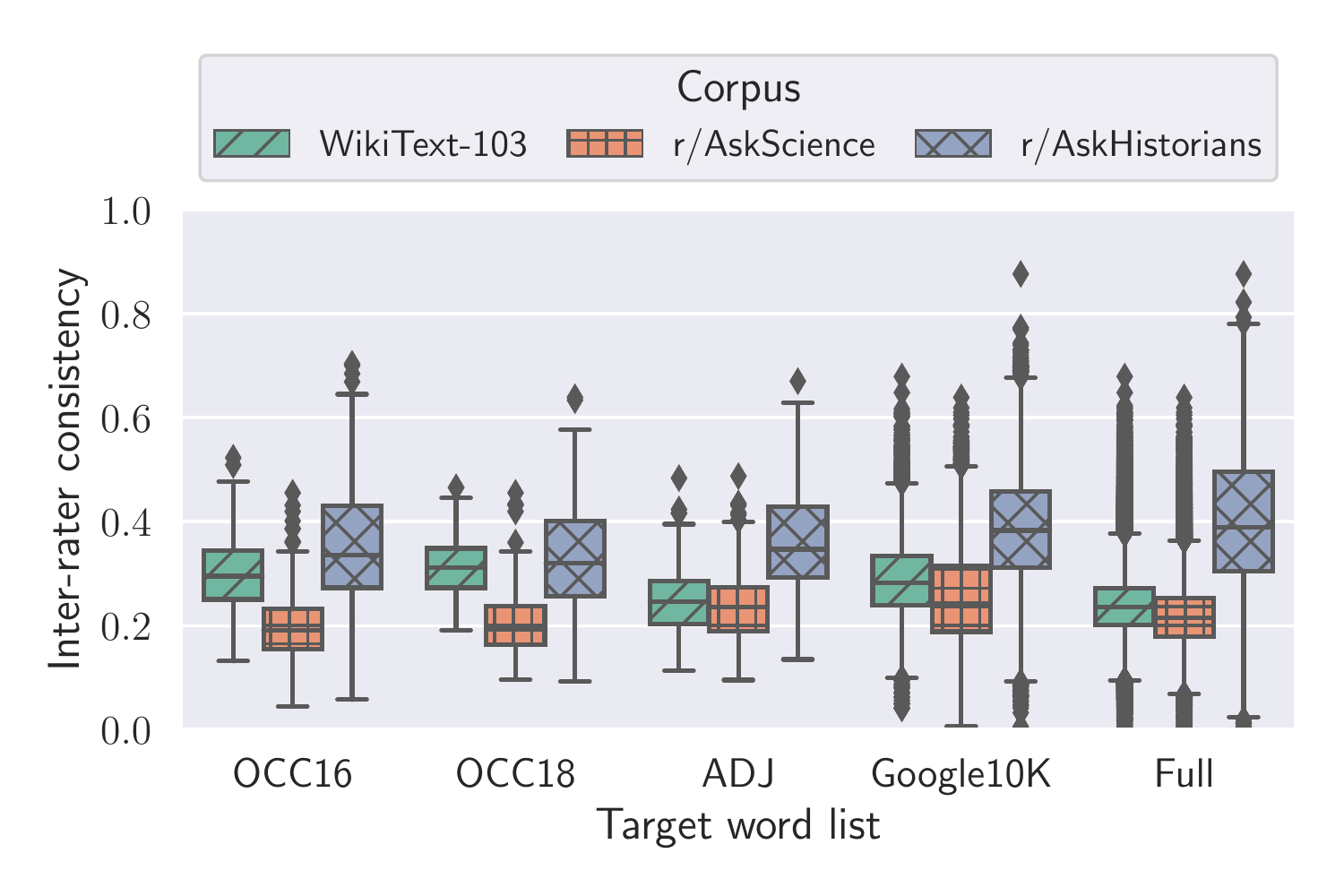}
        \caption{Target words}
        \label{subfig:rater_target}
    \end{subfigure}
    \vfill
    \begin{subfigure}{0.48\textwidth}
        \centering
        \includegraphics[width=\textwidth]{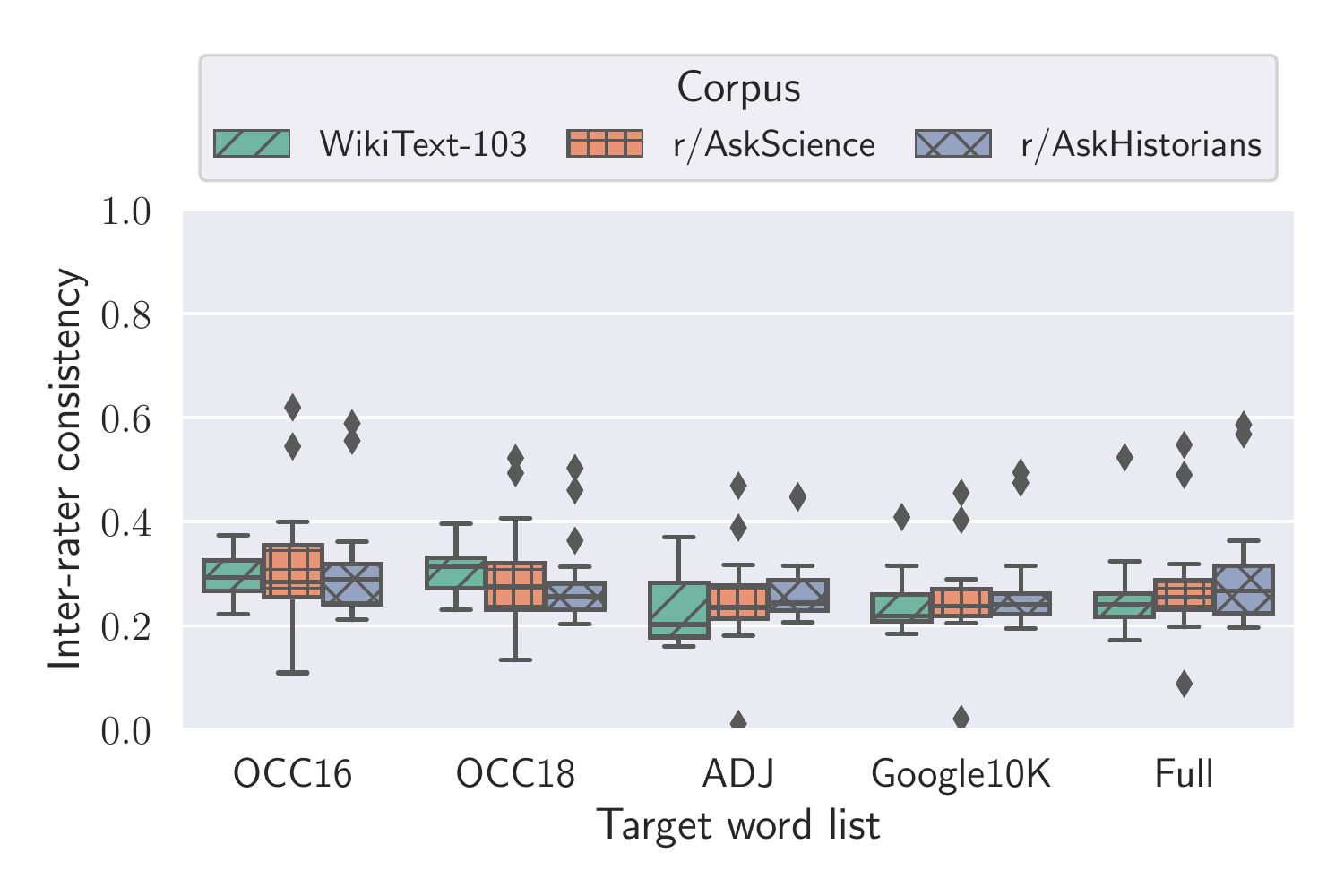}
        \caption{Gender base pairs}
        \label{subfig:rater_gbp}
    \end{subfigure}
    \caption{Distribution of inter-rater consistency scores across target word lists and corpora (based on GloVe). For both target words and gender base pairs, we observe generally low consistency in bias scores across the three scoring rules, regardless of the corpus used.}
    \label{fig:rater_box}
\end{figure}

Figure \ref{fig:rater_box} shows the distributions of inter-rater consistency scores of 
both target words and gender base pairs across different corpora 
(word embeddings trained by GloVe).
More (similar) results on different algorithms 
are in Appendix \ref{app:inter_rater_consistency_figure}. 

We observe that 
the inter-rater consistency of 
the majority of both target words and gender base pairs are rather low. 
This finding 
suggests that different scoring rules
may measure very different aspects of word embedding gender biases, 
and hence their resulting bias scores differ substantially. 
More closely, we observe that for target words, 
the bias scores are the least similar between RIPA and NBM (Pearson’s $r$: 0.836, $p < 0.05$), 
while they are much more similar between DB/WA and RIPA (Pearson’s $r$: 0.923, $p < 0.05$), 
and between DB/WA and NBM (Pearson’s $r$: 0.897, $p < 0.05$). 
A possible reason is that DB/WA and RIPA scores are both based on cosine similarities, and that NBM scores are based on DB/WA scores of the closest neighbours.
In contrast, RIPA and NBM scores are computed in less comparable ways.
Nevertheless, future studies are needed to further investigate the differences 
among scoring rules.

\subsection{Results: Internal Consistency}\label{subsec:internal}

\begin{figure}
    \centering
    \includegraphics[width=0.48\textwidth]{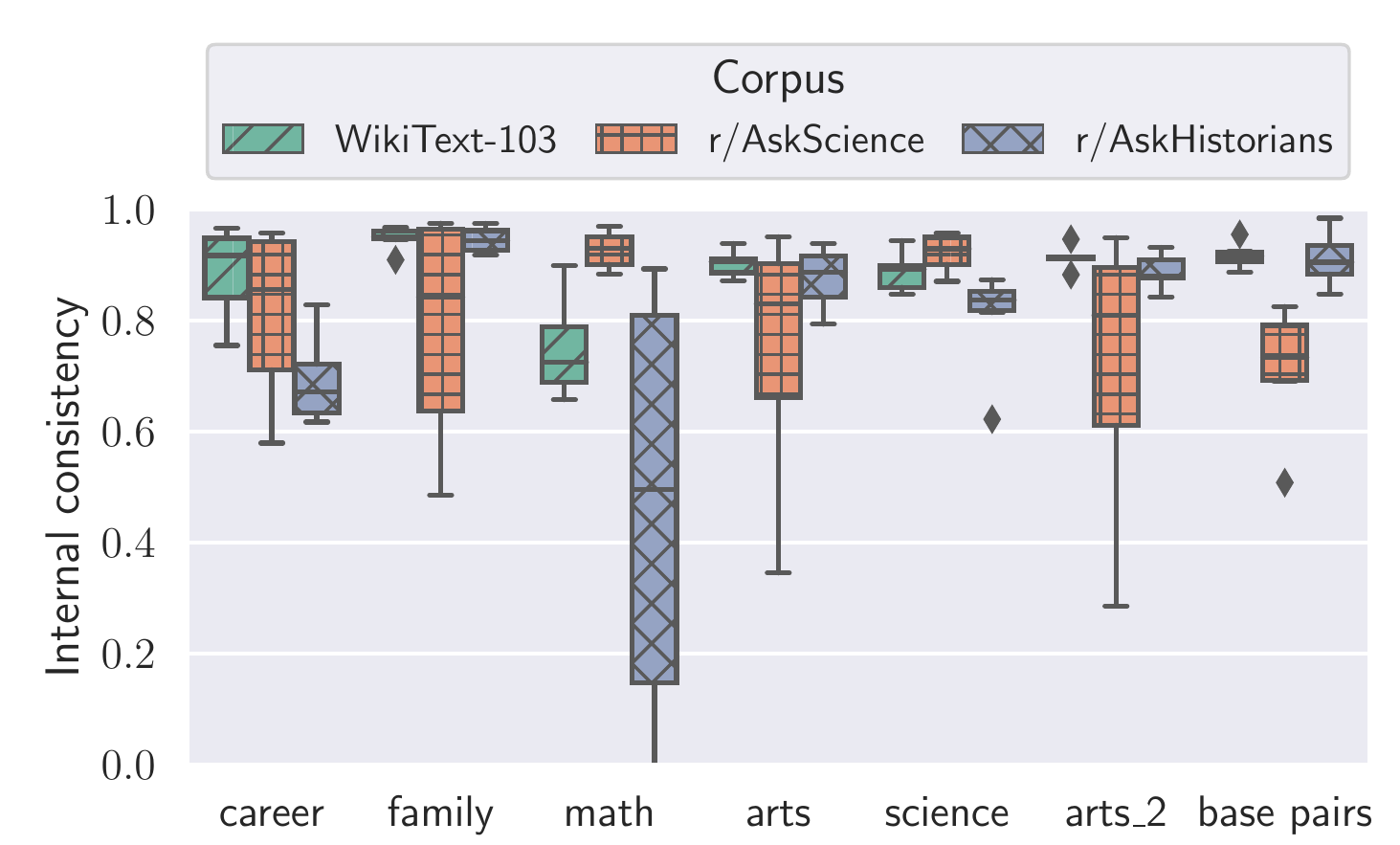}
    \caption{Distribution of internal consistency scores of
    gender bias related queries (e.g., \emph{career}) 
    and the ensemble of gender base pairs (\emph{base pairs}) 
    across corpora.
    Each query and the ensemble of gender base pairs has 
    six reliability scores across different combinations of embedding algorithms and scoring rules.
    While many queries and the ensemble of gender base pairs show 
    overall acceptable internal consistency, 
    the specific scores can still highly depend on 
    the specific query, corpus, scoring rule and even word embedding algorithm used.}
    \label{fig:internal}
\end{figure}

Figure~\ref{fig:internal} presents the distribution of the internal consistency scores of 
every query 
and the ensemble of all gender base pairs across corpora. 
Each boxplot contains six scores from the combinations of embedding algorithms and scoring rules. 
We make four observations: 

First, the internal consistency of most queries 
and the ensemble of gender base pairs are acceptable (Cronbach's alpha values $\geq$ 0.7).\footnote{
Cronbach's alpha values greater than 0.7 are considered acceptable. See \citet{cicchetti_guidelines_1994}.
} 
This indicates that 
most target words in the same query likely
measure gender bias of the same concept.
Bias scores of a target word are also generally consistent across gender base pairs.

Second, however, the patterns of internal consistency vary substantially across queries. 
For example, on the \emph{WikiText-103} corpus, 
the internal consistency scores of \emph{family} are much higher and less varied than the scores of \emph{math}.

Third, the internal consistency of a query and the ensemble of gender base pairs seems dependent on specific corpora.
For instance, the internal consistency scores of 
\emph{math} are high and have a low variance on the corpus \emph{r/AskScience}, 
but they are low and have a very high variance on \emph{r/AskHistorians}.

Fourth, 
the high variance of scores for some queries (e.g. \emph{math} on \emph{r/AskHistorians}) suggests that
a query's internal consistency may depend also on word embedding algorithms and scoring rules.

\subsection{Factors Influencing 
the Reliability of Gender Base Pairs and Target Words}\label{subsec:factors}

In this section, 
we investigate 
factors influencing the test-retest reliability and inter-rater consistency of 
both gender base pairs and target words. 
Because we only have a small number of gender base pairs, we qualitatively inspect them using visualisations.
For (the large number of) target words, 
we use regression to model the effects of the factors.

\begin{figure}[t]
    \centering
    \begin{subfigure}{0.48\textwidth}
        \centering
        \includegraphics[width=\textwidth]{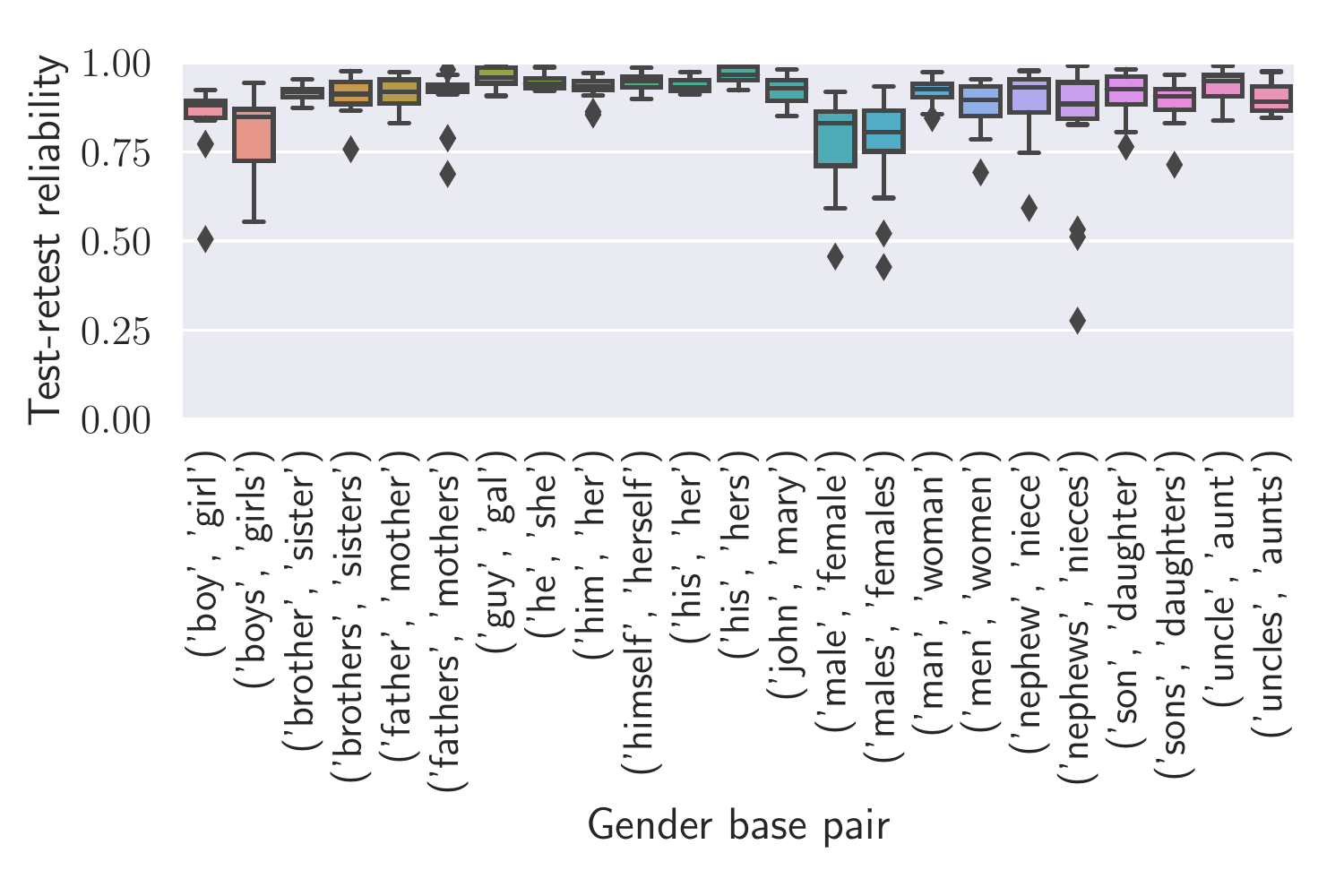}
        \caption{Test-retest reliability}
        \label{fig:retest_box_gbp}
    \end{subfigure}
    \begin{subfigure}{0.48\textwidth}
        \centering
        \includegraphics[width=\textwidth]{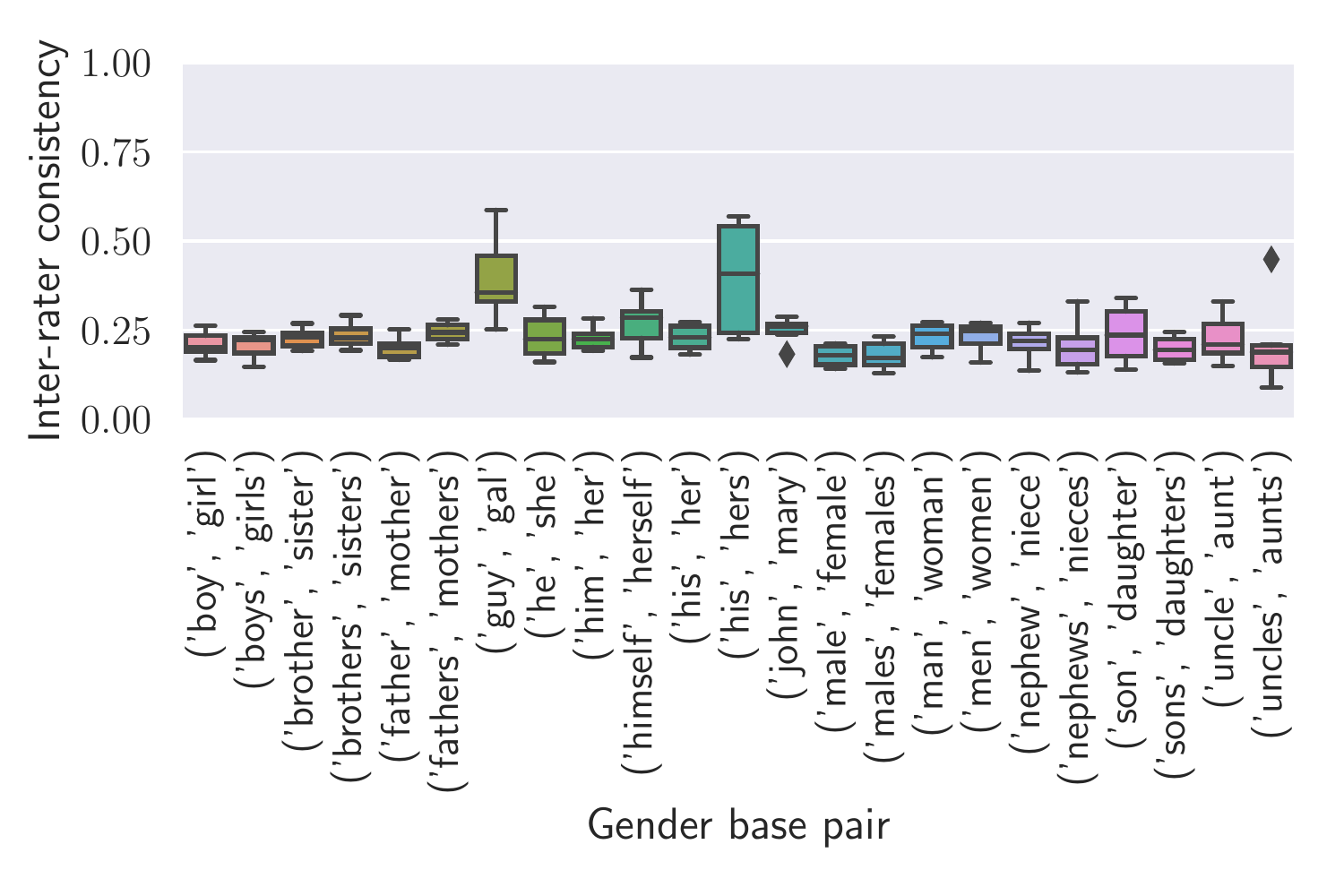}
        \caption{Inter-rater consistency}
        \label{fig:rater_box_gbp}
    \end{subfigure}
    \caption{Test-retest reliability and inter-rater consistency 
    of different gender base pairs on full vocabularies.
    Each gender base pair has multiple reliability scores across combinations of embedding algorithms and corpora (as well as scoring rules for test-retest reliability).
    Gender base pairs in singular form tend to have higher test-retest reliability.
    Also, gender base pairs with higher test-retest reliability are more likely to score higher in inter-rater consistency.}
    \label{fig:box_gbp}
\end{figure}

\paragraph{Gender Base Pairs: Visualisation}

The distributions of test-retest reliability and 
inter-rater consistency of gender base pairs 
(on full vocabularies) 
are shown in Figure~\ref{fig:box_gbp}. 
We make two observations: 

First, gender base pairs in singular form are usually of higher test-retest reliability 
(e.g. \emph{boy}$\sim$\emph{girl} versus
\emph{boys}$\sim$\emph{girls}), 
which is consistent with findings by \citet{zhang2020robustness}. 
The median difference in test-retest reliability between singular and plural gender base pairs 
is statistically significant ($t(7) = 2.45, p < .05$).
In contrast, such a statistical difference is not found 
for inter-rater consistency ($t(7) = -0.13, p > .05$). 
        
Second, 
gender base pairs of higher test-retest reliability 
also tend to be of higher inter-rater consistency, 
evidenced by the moderate correlation between the median test-retest reliability scores 
and the median inter-rater consistency scores of gender base pairs
($r=0.644$, $p<.05$).

\paragraph{Target Words: Regression Analyses}

\begin{table}[t]
\centering
\small
\begin{tabular}{c|cc|cc}
\toprule
            & \multicolumn{2}{c}{Test-retest} & \multicolumn{2}{c}{Inter-rater}\\ 
            & Estimate & $\Delta R^2$ & Estimate & $\Delta R^2$\\ \midrule
$\text{SR}_\text{DB/WA}$ &  reference & 0.0040 & - & -\\
$\text{SR}_\text{RIPA}$ &  \textbf{-0.0102} & - & - & -\\
$\text{SR}_\text{NBM}$ &  -0.0001 & - & - & -\\
log freq         & \textbf{0.0062} & 0.0275 & \textbf{0.0241} & 0\\
log$^2$ freq       & \textbf{0.0023} & 0.0051 & \textbf{0.0008} & 0\\
log \#senses & \textbf{-0.0012} & 0.0002  & \textbf{-0.0007} & 0\\
$\text{PoS}$  & - & 0.0093 & - & 0.0022\\
NN Sim & \textbf{-0.0024} & 0.0120 &  \textbf{0.0011}& 0.0038\\
L2 norm & \textbf{-0.0118} & 0.0456 & \textbf{-0.0051}& 0\\
ES & \textbf{0.0457} & 0.1020 &  \textbf{0.0186}&0\\ \midrule
$R^2_{\text{fixed}}$& 0.3261 & - &0.0581&-\\
$R^2_{\text{corpus}}$& 0.0113 & - &0.0271&-\\
$R^2_{\text{algorithm}}$& 0.1802 & - &0.4307 &-\\
$R^2_{\text{total}}$ & 0.5177 & - &0.5160 &-\\ \bottomrule
\end{tabular}
\caption{Results of multilevel regression on 
    the test-retest and inter-rater reliability of target words.
    Estimates are standardised (bold if $p < 0.05$).
    SR: scoring rule. 
    $\Delta R^2$ is the reduction in explained variance 
    when a factor is left out.
    $R^2_{\text{fixed}}$, $R^2_{\text{corpus}}$, $R^2_{\text{algorithm}}$ 
    and $R^2_{\text{total}}$ 
    refer to the explained variance of fixed factors 
    (i.e. word-level features and scoring rules), 
    embedding training corpora, 
    embedding training algorithms, 
    total effects of all these three parts, 
    respectively. 
}
\label{tab:multilevel}
\end{table}

We use multilevel regression to study potential 
influencing factors of
the test-retest reliability and inter-rater consistency 
of target words.\footnote{
    We offer a more detailed description of multilevel regression, 
    the used features 
    and PoS estimates
    in Appendix \ref{app:mlr}.}
Comparing with 
OLS regression and its variants, 
multilevel models  
allow for dependent observations. 
Therefore, they suit our data better where
reliability scores are nested within groups 
(e.g. different training algorithms and corpora of embeddings) 
and are thus correlated.
Multilevel models have a further advantage that 
they estimate not only the effects of fixed factors 
(i.e. standard features) 
but also the amount of variance explained by each grouping factor. 

We collect a range of word-level features as fixed factors, 
mostly inspired by previous studies 
\citep{burdick2018factors,pierrejean-tanguy-2018-predicting,DBLP:conf/coling/HellrichH16}.
These include 1) word-intrinsic features: log number of WordNet synsets ($\log$ \#senses) 
and the most common Part-of-Speech tag (PoS) in the Brown corpus \citep{francis1979brown}, 
as in \citet{burdick2018factors}; 
2) corpus-related features: 
log frequencies of words in the training corpus ($\log$ freq) 
and their squares ($\log^2\text{freq}$); 
3) embedding-related features: 
cosine similarity to nearest neighbour (NN Sim), 
L2 norm (L2 Norm) 
and embedding stability (ES). 
We calculate ES as follows: 
for each pair of word embedding models, 
we first fit an orthogonal transformation $Q$ that minimizes the Frobenius norm of their difference. 
The stability of a word across multiple random seeds is then 
calculated as the average pairwise cosine similarity of its embedding vectors after transformation by $Q$s.   
We also consider scoring rules as a fixed factor 
because we are interested in 
comparing the influence of these three scoring rules 
on target words' test-retest reliability.
The two grouping factors are embedding algorithms and training corpora.

We summarise the results in Table \ref{tab:multilevel}. 
For test-retest reliability, the model has a satisfactory
total explained variance ($R^2_{\text{model}}: 51.77\%$). 
Fixed factors (including scoring rules) together 
explain a substantial part of the variation ($R^2_{\text{fixed}}: 32.61\%$). 
Among these factors, embedding stability (ES) appears to be 
the most important one, 
indicated by the largest standardised effect estimate and $\Delta R^2$. 
The higher the embedding stability, the higher the test-retest reliability, which is expected.
L2 norm and word frequency 
also account for a considerable amount of variance. 
When the L2 norm is lower or when the frequency is higher, 
test-retest reliability is higher.
This observation is also consistent with prior research findings. 
For instance, \citet{DBLP:conf/coling/HellrichH16} show that word frequency positively correlates 
with embedding stability when word frequency is not too high. 
Also, \citet{arora-etal-2016-latent} find that 
the L2 norm correlates negatively with word frequency. 
This finding 
agrees with our observation in Figure \ref{fig:retest_box} as well.
In contrast, the choice of scoring rules has only a minor impact on test-retest reliability ($R^2: 0.4\%$).
Despite a statistically significant difference between DB/WA and RIPA, the difference is very small (-0.0102) and therefore unlikely important. 

Among the group level factors, 
embedding algorithms alone 
explain 18.02\% of the total variance. 
This suggests that 
the test-retest reliability of a target word 
is determined 
by the word embedding training to a considerable degree.
In contrast, the choice of corpora is much less influential.

For inter-rater consistency, 
the resulting model is also of good explanatory power ($R^2_{\text{model}}: 51.60\%$). 
However, 
it is clear that word-level features 
fail to explain much of the variance (5.81\%). 
Between the two grouping factors, algorithm dominates with an $R^2$ score of 0.4307. 
This indicates that the inter-rater reliability of a target word 
is largely determined by the word embedding algorithm used. 

Note that we also explored potential interactions between the fixed factors and 
how they might impact the outcome test-retest and inter-rater reliability scores. 
However, it turned out that interaction effects 
generally had small effect sizes and did not considerably improve overall model fit.
We therefore excluded them from the final models.

\section{Conclusion \& Discussion}
In this paper, 
we propose to leverage measurement theory to examine 
the reliability of word embedding bias measures. 
We find that 
bias scores are mostly consistent across different random seeds (i.e. high test-retest reliability), 
as well as across
gender base pairs and 
target words within a query (i.e. high internal consistency).  
In contrast, the three scoring rules fail to agree with one another 
(i.e. low inter-rater consistency).
Furthermore, our regression results suggest that 
the consistency of bias scores 
across different random seeds 
are mostly influenced by various word-level features 
as well as the word embedding algorithm used. 
Meanwhile, the bias scores of target words across different scoring rules 
are dominated by the word embedding algorithm used.
We thus urge future studies to be more critical about 
applying such measures. 

Nevertheless, our work has limitations.
First, we only consider gender bias measures.
Future work should apply our reliability evaluation framework to other types of bias (e.g. racial bias). 
Second, we focus on static word embeddings. 
Future work should investigate the reliability of bias measures for contextualised embeddings. 
Third, 
we do not address
validity, the other crucial aspect of measurement quality. 
We thus call for future studies on 
the validity of word embedding bias measures.
Fourth, \citet{goldfarb-tarrant-etal-2021-intrinsic} argue that 
intrinsic (word embeddings) biases sometimes fail to agree with
extrinsic biases (measured in downstream tasks, e.g. coreference resolution). 
One potential research direction is to assess the reliability 
of extrinsic bias measurements as well, 
to shed further light on the disconnect between intrinsic and extrinsic biases.
Lastly, while ICC and Cronbach's Alpha are established reliability estimators in many scientific disciplines,
correct interpretation of their values is often challenging
and requires both statistical and field-specific expertise \citep{lee_pitfalls_2012, streiner_starting_2003}.
Future work should address the appropriate use of these estimators and their limitations in the context of NLP research.

\section*{Acknowledgements}

We thank all anonymous reviewers for their constructive and helpful feedback. 
We also thank Anna Wegmann for the proofreading and productive discussions.  
This work was partially supported by the Dutch Research Council (NWO) (grant number: VI.Veni.192.130 to D. Nguyen; grant number: VI.Vidi.195.152 to D. L. Oberski).

\section*{Ethical Statement}

\paragraph{Intended Usage}
As aforementioned in \S \ref{sec:intro}, 
word embedding bias measures are often used 
to analyse word embedding models, to assess the effect of bias mitigation methods,
and to study societal biases.
Our work thus intends to evaluate the quality of these measures 
and their derived conclusions. 
Moreover, our framework can also be used to assess the reliability of bias measures 
which consists of target words, gender base pairs, and scoring rules 
that were not included in this study. 
In this way, our framework can contribute to the development of models that are less biased and hence potentially less harmful.

\paragraph{Limitations}
In this study, we focused on common measures of gender biases in word embeddings. 
Measurements of gender biases in word embeddings typically rely on manually crafted sets of target words and pairs of gendered words (i.e. gender base pairs, such as \textit{he} vs. \textit{she}).
In our experiments we use existing lists of words and word pairs that have been frequently used in related work. However, these word pairs were constructed by taking the very narrow view of binary gender.
We hope to see more work on measures of bias in embeddings that considers non-binary gender identities as well as intersectional identities.

\bibliographystyle{acl_natbib}
\bibliography{emnlp2021}

\clearpage
\appendix
\section{Word Lists}

\subsection{Gender Base Pairs 
\citep{bolukbasi2016man,caliskan2017semantics,garg2018word}.}\label{app: gbp}

boy $\sim$ girl, boys $\sim$ girls, brother $\sim$ sister, 
brothers $\sim$ sisters, father $\sim$ mother, fathers $\sim$ mothers, 
guy $\sim$ gal, he $\sim$ she, him $\sim$ her, himself $\sim$ herself, 
his $\sim$ her, his $\sim$ hers, john $\sim$ mary, male $\sim$ female, 
males $\sim$ females, man $\sim$ woman, men $\sim$ women, 
nephew $\sim$ niece, nephews $\sim$ nieces, son $\sim$ daughter, 
sons $\sim$ daughters, uncle $\sim$ aunt, uncles $\sim$ aunts.

\subsection{Target Word Lists}

\paragraph{OCC16 \citep{bolukbasi2016man}}
accountant, acquaintance, actor, actress, adjunct\_professor, administrator, adventurer, advocate,
aide, alderman, alter\_ego, ambassador, analyst, anthropologist, archaeologist, archbishop,
architect, artist, artiste, assassin, assistant\_professor, associate\_dean, associate\_professor, astronaut,
astronomer, athlete, athletic\_director, attorney, author, baker, ballerina, ballplayer,
banker, barber, baron, barrister, bartender, biologist, bishop, bodyguard,
bookkeeper, boss, boxer, broadcaster, broker, bureaucrat, businessman, businesswoman,
butcher, butler, cab\_driver, cabbie, cameraman, campaigner, captain, cardiologist,
caretaker, carpenter, cartoonist, cellist, chancellor, chaplain, character, chef,
chemist, choreographer, cinematographer, citizen, civil\_servant, cleric, clerk, coach,
collector, colonel, columnist, comedian, comic, commander, commentator, commissioner,
composer, conductor, confesses, congressman, constable, consultant, cop, correspondent,
councilman, councilor, counselor, critic, crooner, crusader, curator, custodian,
dad, dancer, dean, dentist, deputy, dermatologist, detective, diplomat,
director, disc\_jockey, doctor, doctoral\_student, drug\_addict, drummer, economics\_professor, economist,
editor, educator, electrician, employee, entertainer, entrepreneur, environmentalist, envoy,
epidemiologist, evangelist, farmer, fashion\_designer, fighter\_pilot, filmmaker, financier, firebrand,
firefighter, fireman, fisherman, footballer, foreman, freelance\_writer, gangster, gardener,
geologist, goalkeeper, graphic\_designer, guidance\_counselor, guitarist, hairdresser, handyman, headmaster,
historian, hitman, homemaker, hooker, housekeeper, housewife, illustrator, industrialist,
infielder, inspector, instructor, interior\_designer, inventor, investigator, investment\_banker, janitor,
jeweler, journalist, judge, jurist, laborer, landlord, lawmaker, lawyer,
lecturer, legislator, librarian, lieutenant, lifeguard, lyricist, maestro, magician,
magistrate, maid, major\_leaguer, manager, marksman, marshal, mathematician, mechanic,
mediator, medic, midfielder, minister, missionary, mobster, monk, musician,
nanny, narrator, naturalist, negotiator, neurologist, neurosurgeon, novelist, nun,
nurse, observer, officer, organist, painter, paralegal, parishioner, parliamentarian,
pastor, pathologist, patrolman, pediatrician, performer, pharmacist, philanthropist, philosopher,
photographer, photojournalist, physician, physicist, pianist, planner, plastic\_surgeon, playwright,
plumber, poet, policeman, politician, pollster, preacher, president, priest,
principal, prisoner, professor, professor\_emeritus, programmer, promoter, proprietor, prosecutor,
protagonist, protege, protester, provost, psychiatrist, psychologist, publicist, pundit,
rabbi, radiologist, ranger, realtor, receptionist, registered\_nurse, researcher, restaurateur,
sailor, saint, salesman, saxophonist, scholar, scientist, screenwriter, sculptor,
secretary, senator, sergeant, servant, serviceman, sheriff\_deputy, shopkeeper, singer,
singer\_songwriter, skipper, socialite, sociologist, soft\_spoken, soldier, solicitor, solicitor\_general,
soloist, sportsman, sportswriter, statesman, steward, stockbroker, strategist, student,
stylist, substitute, superintendent, surgeon, surveyor, swimmer, taxi\_driver, teacher,
technician, teenager, therapist, trader, treasurer, trooper, trucker, trumpeter,
tutor, tycoon, undersecretary, understudy, valedictorian, vice\_chancellor, violinist, vocalist,
waiter, waitress, warden, warrior, welder, worker, wrestler, write.

\paragraph{OCC18 \citep{garg2018word}}
janitor, statistician, midwife, bailiff, auctioneer,
photographer, geologist, shoemaker, athlete, cashier,
dancer, housekeeper, accountant, physicist, gardener,
dentist, weaver, blacksmith, psychologist, supervisor,
mathematician, surveyor, tailor, designer, economist,
mechanic, laborer, postmaster, broker, chemist,
librarian, attendant, clerical, musician, porter,
scientist, carpenter, sailor, instructor, sheriff,
pilot, inspector, mason, baker, administrator,
architect, collector, operator, surgeon, driver,
painter, conductor, nurse, cook, engineer,
retired, sales, lawyer, clergy, physician,
farmer, clerk, manager, guard, artist, smith, official, 
police, doctor, professor, student, judge, teacher,
author, secretary, soldier.

\paragraph{ADJ \citep{garg2018word}}
headstrong, thankless, tactful, distrustful, quarrelsome, effeminate, fickle, talkative, dependable, resentful,
sarcastic, unassuming, changeable, resourceful, persevering, forgiving, assertive, individualistic, vindictive, sophisticated,
deceitful, impulsive, sociable, methodical, idealistic, thrifty, outgoing, intolerant, autocratic, conceited,
inventive, dreamy, appreciative, forgetful, forceful, submissive, pessimistic, versatile, adaptable, reflective,
inhibited, outspoken, quitting, unselfish, immature, painstaking, leisurely, infantile, sly, praising,
cynical, irresponsible, arrogant, obliging, unkind, wary, greedy, obnoxious, irritable, discreet,
frivolous, cowardly, rebellious, adventurous, enterprising, unscrupulous, poised, moody, unfriendly, optimistic,
disorderly, peaceable, considerate, humorous, worrying, preoccupied, trusting, mischievous, robust, superstitious,
noisy, tolerant, realistic, masculine, witty, informal, prejudiced, reckless, jolly, courageous,
meek, stubborn, aloof, sentimental, complaining, unaffected, cooperative, unstable, feminine, timid,
retiring, relaxed, imaginative, shrewd, conscientious, industrious, hasty, commonplace, lazy, gloomy,
thoughtful, dignified, wholesome, affectionate, aggressive, awkward, energetic, tough, shy, queer,
careless, restless, cautious, polished, tense, suspicious, dissatisfied, ingenious, fearful, daring,
persistent, demanding, impatient, contented, selfish, rude, spontaneous, conventional, cheerful, enthusiastic,
modest, ambitious, alert, defensive, mature, coarse, charming, clever, shallow, deliberate,
stern, emotional, rigid, mild, cruel, artistic, hurried, sympathetic, dull, civilized,
loyal, withdrawn, confident, indifferent, conservative, foolish, moderate, handsome, helpful, gentle,
dominant, hostile, generous, reliable, sincere, precise, calm, healthy, attractive, progressive,
confused, rational, stable, bitter, sensitive, initiative, loud, thorough, logical, intelligent,
steady, formal, complicated, cool, curious, reserved, silent, honest, quick, friendly,
efficient, pleasant, severe, peculiar, quiet, weak, anxious, nervous, warm, slow,
dependent, wise, organized, affected, reasonable, capable, active, independent, patient, practical,
serious, understanding, cold, responsible, simple, original, strong, determined, natural, kind.

\subsection{Queries \citep{caliskan2017semantics}}

\paragraph{Career} 
  executive, management, professional, corporation, salary, office, business, career.

\paragraph{Family} 
  home, parents, children, family, cousins, marriage, wedding, relatives.

\paragraph{Arts}
  poetry, art, dance, literature, novel, symphony, drama, sculpture.

\paragraph{Arts\_2}
  poetry, art, shakespeare, dance, literature, novel, symphony, drama.

\paragraph{Math}
  math, algebra, geometry, calculus, equations, computation, numbers, addition.

\paragraph{Science}
  science, technology, physics, chemistry, einstein, nasa, experiment, astronomy.

\section{Environmental Setup}
\paragraph{Running Environments}
All experiments except for multilevel modelling were performed on 
Intel Xeon E5-2699 CPUs with Python 3.7.
Running these experiments took approximately 700 CPU hours. 
Training the word embedding models with different random seeds  took up most of the time (500 CPU hours). 
Moreover, calculating NBM bias scores for the full vocabularies 
was computationally expensive (more than 100 CPU hours). 
However, fortunately, the required consumption 
has a linear relationship with the number of target words. 
Thus, considerably fewer computational resources
will be needed to estimate the reliability 
of a small set of target words. 

Multilevel modelling was conducted 
in the statistical software R (version: 4.0.1) 
on an Intel Core i7-8565U CPU, 
consuming approximately 3 CPU hours. 

\paragraph{Data Preprocessing}
For Reddit data (i.e. \emph{r/AskScience} and \emph{r/AskHistorians}), 
we lower-cased, removed redundant spaces/urls, and  
used the Spacy\footnote{\url{https://github.com/explosion/spaCy}} library 
to tokenize each sentence. 
For training GloVe embeddings, we substituted ``<unk>'' symbols in 
WikiText-103 with ``<raw\_unk>'' symbols. 

\section{Choosing ICC Estimators}\label{app:ICC}
Despite ICC being a commonly used tool for estimating test-retest and inter-rater reliability, 
there exist distinct forms of ICC estimators.
Different forms of ICC can involve distinct assumptions and can therefore lead to very different interpretations \cite{koo2016guideline}.

In this work, we follow the framework proposed by \citet{shrout_intraclass_1979}.
They define 6 forms of ICC and present them as ``ICC'' with 2 additional numbers in parentheses (e.g., ICC(2,1) and ICC(3,1)). 
The first number refers to the \textbf{model} and can take on three possible values (1, 2 or 3):
1 is a one-way random-effects model, where each subject receives a unique, random set of raters; 
2 is a two-way random-effects model, where all subjects receive the same randomly chosen set of raters and the reliability results are assumed to be generalisable to unseen raters;
3 is a two-way mixed-effects model, where the selected raters are the only raters of interest and thus the reliability results are not generalisable to other raters. 
The second number refers to the \textbf{intended use} of raters/measurements in an application and can take on two values (1 or k).
1 refers to having only a single rater or measurement; 
k means using the mean of k raters or measurements.

Therefore, depending on the specific research data and goals, one of the 6 ICC forms may be used. 
For test-retest reliability, we use ICC(2,1) for the following three reasons. 
First, the raters (i.e. different random seeds) are a random sample of the population (of all possible random seeds).
Second, each bias score receives the same raters (i.e. random seeds).
Third, in actual research practices, researchers would normally use only one rater (one random seed) to measure word embedding biases.

For inter-rater consistency, we use ICC(3,1) based on two considerations.
First, we are only interested in comparing three specific scoring rules (i.e. raters).
Second, in practice, researchers would use only the result from one scoring rule (i.e. rater) to measure word embedding biases. 

\clearpage

\section{Additional Figures}

\subsection{Test-retest Reliability}\label{app:test_retest_reliability_figure}

\begin{figure}[h!]
    \centering
    \includegraphics[width=0.48\textwidth]{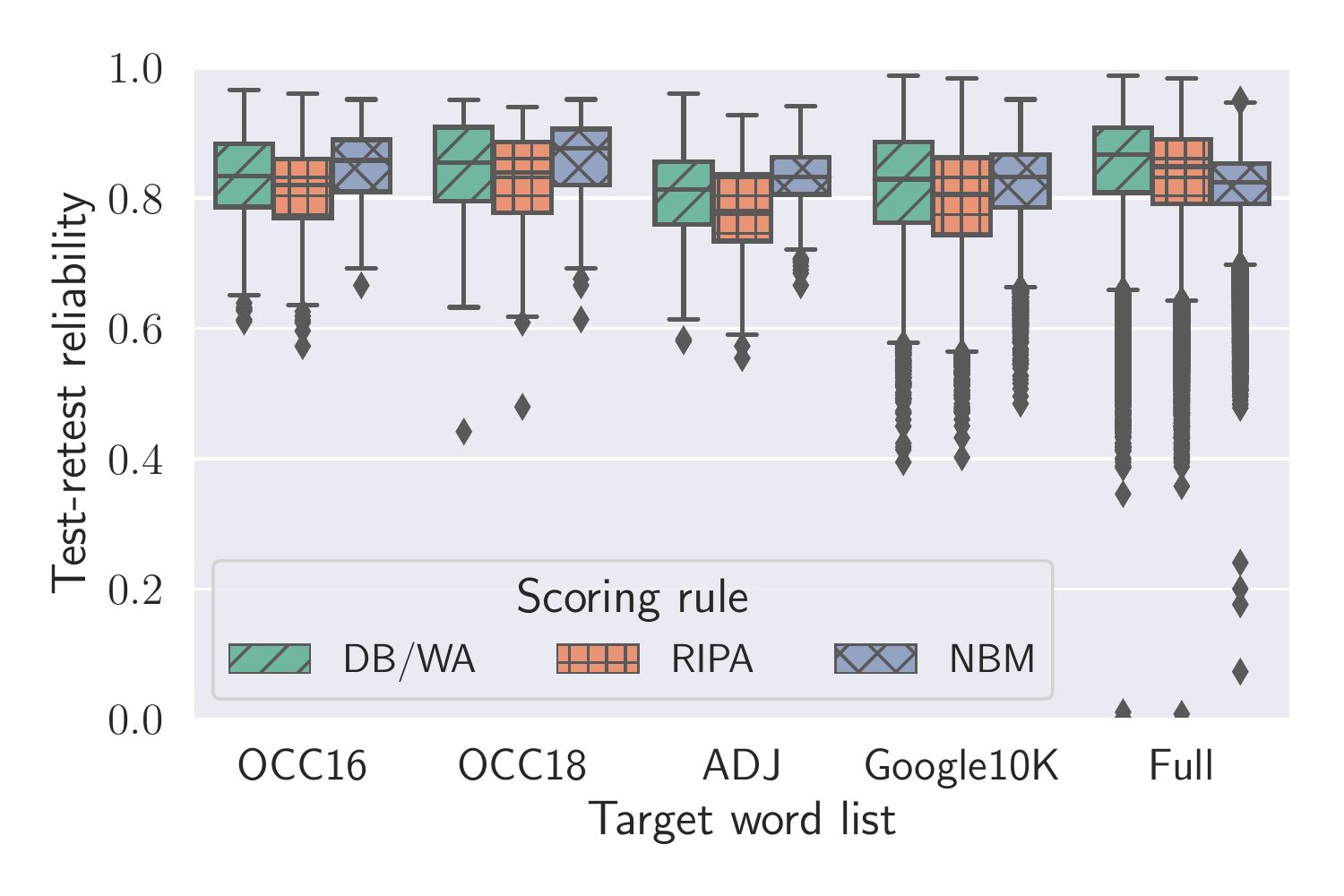}
    \caption{Test-retest reliability of target words. 
     The word embeddings are trained with GloVe on WikiText-103.}
\end{figure}

\begin{figure}[h!]
    \centering
    \includegraphics[width=0.48\textwidth]{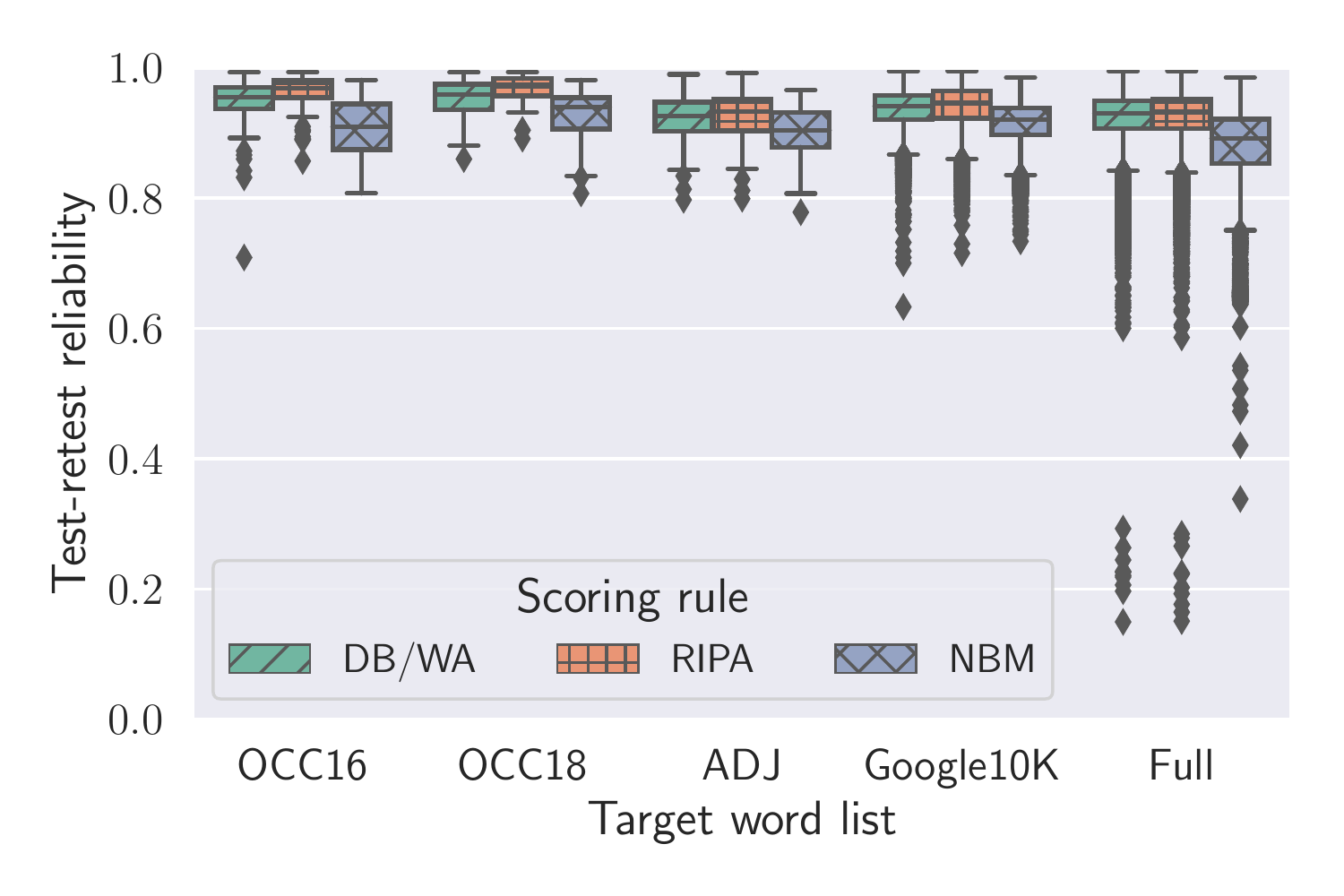}
    \caption{Test-retest reliability of target words. 
     The word embeddings are trained with SGNS on r/AskScience.}
\end{figure}

\begin{figure}[h!]
    \centering
    \includegraphics[width=0.48\textwidth]{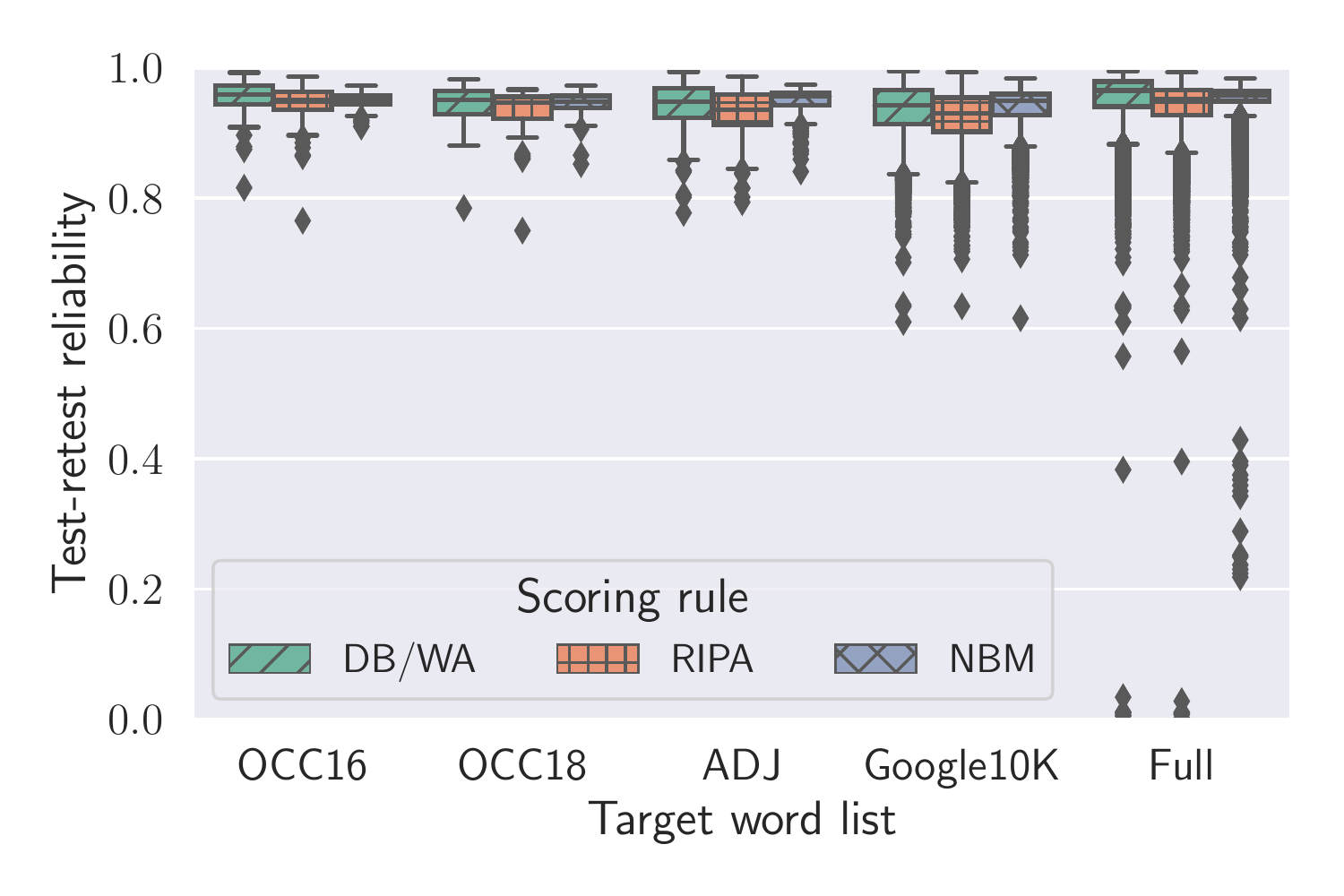}
    \caption{Test-retest reliability of target words.
      The word embeddings are trained with GloVe on r/AskScience.}
\end{figure}

\begin{figure}[h!]
    \centering
    \includegraphics[width=0.48\textwidth]{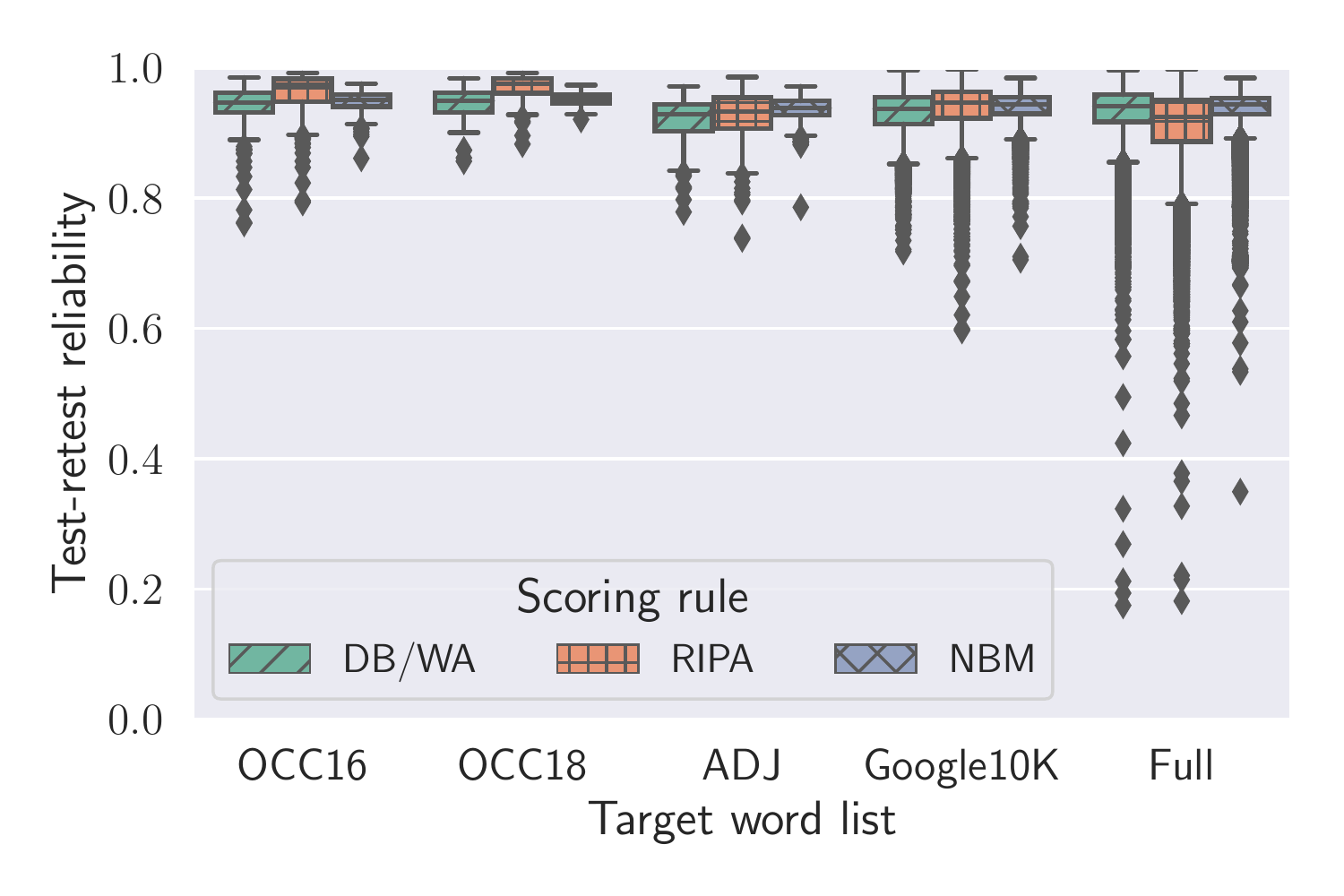}
    \caption{Test-retest reliability of target words. 
     The word embeddings are trained with SGNS on r/AskHistorians.}
\end{figure}

\begin{figure}[h!]
    \centering
    \includegraphics[width=0.48\textwidth]{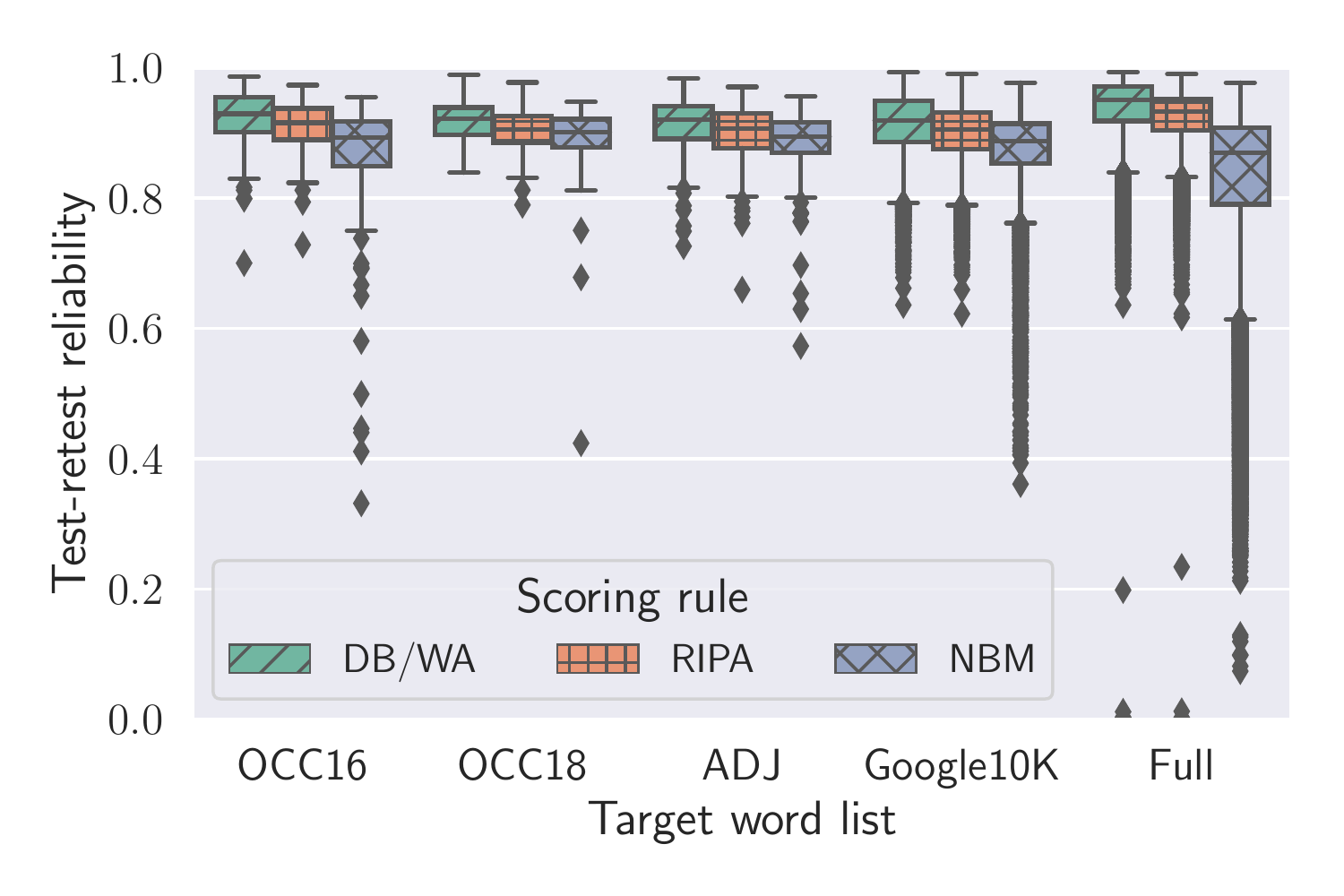}
    \caption{Test-retest reliability of target words, 
     The word embeddings are trained with GloVe on r/AskHistorians.}
\end{figure}

\begin{figure}[h!]
    \centering
    \includegraphics[width=0.48\textwidth]{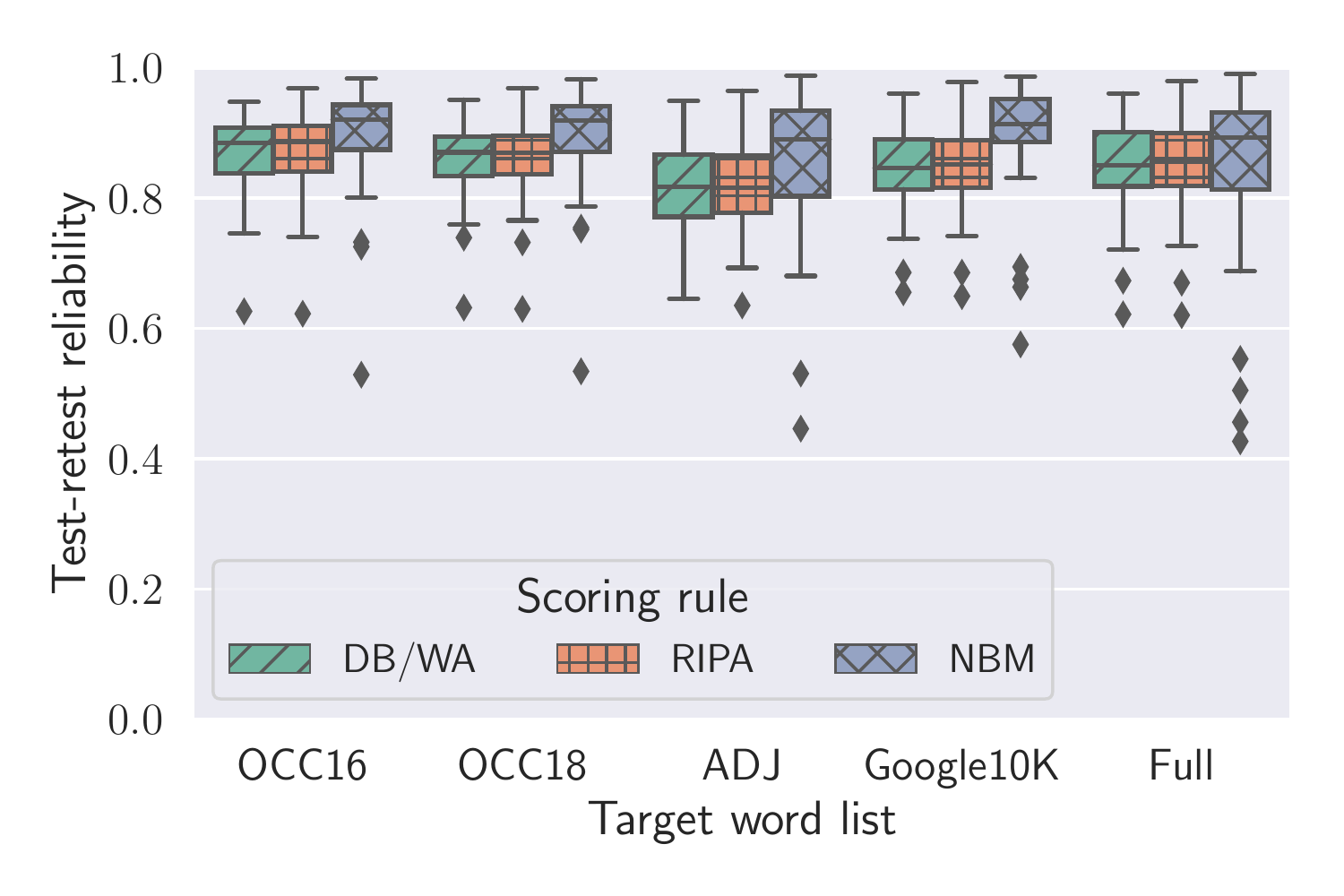}
    \caption{Test-retest reliability of gender base pairs. 
     The word embeddings are trained with GloVe on WikiText-103.}
\end{figure}

\begin{figure}[h!]
    \centering
    \includegraphics[width=0.48\textwidth]{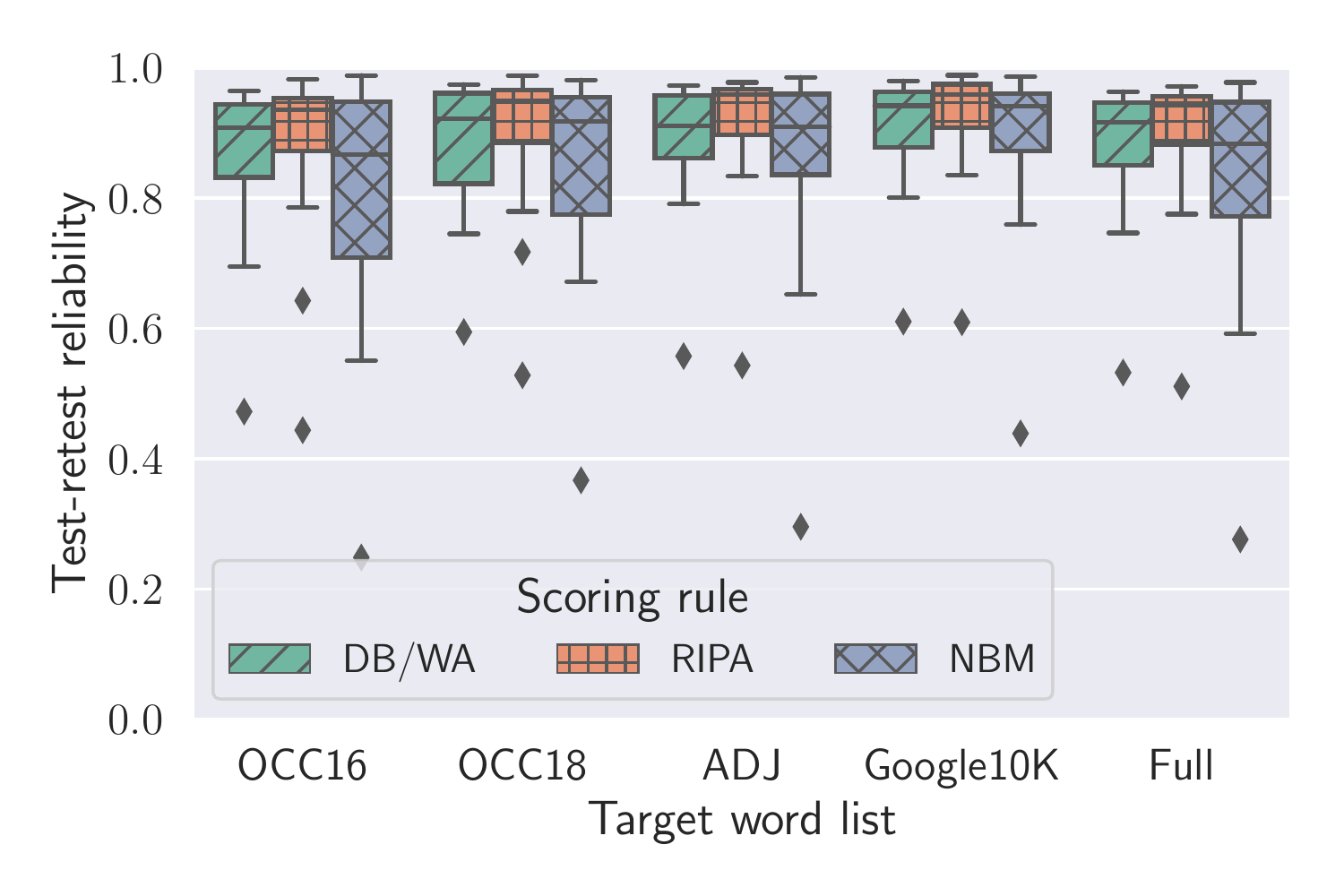}
    \caption{Test-retest reliability of gender base pairs. 
     The word embeddings are trained with SGNS on r/AskScience.}
\end{figure}

\begin{figure}[h!]
    \centering
    \includegraphics[width=0.48\textwidth]{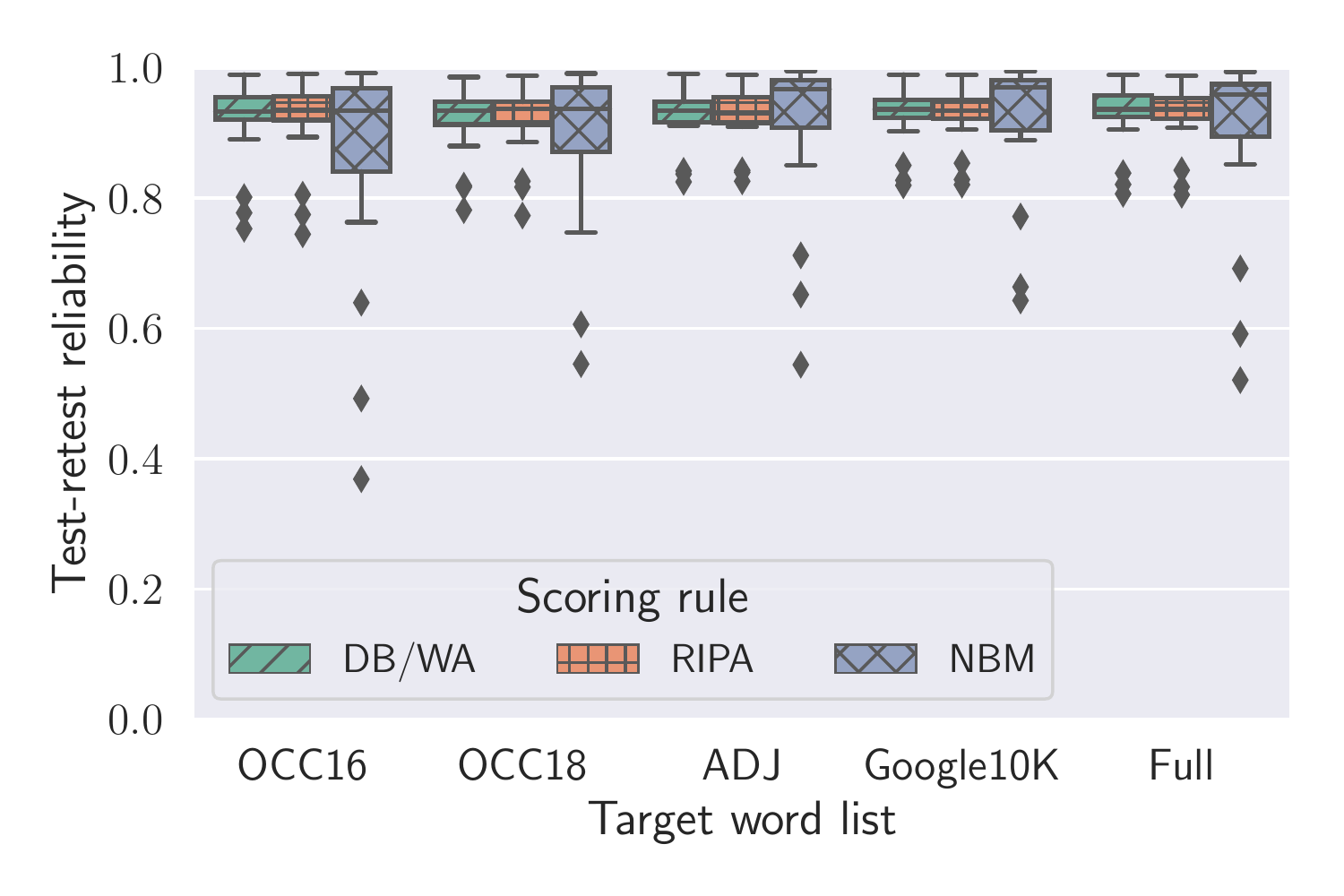}
    \caption{Test-retest reliability of gender base pairs. 
    The  word embeddings are trained with GloVe on r/AskScience.}
\end{figure}

\begin{figure}[h!]
    \centering
    \includegraphics[width=0.48\textwidth]{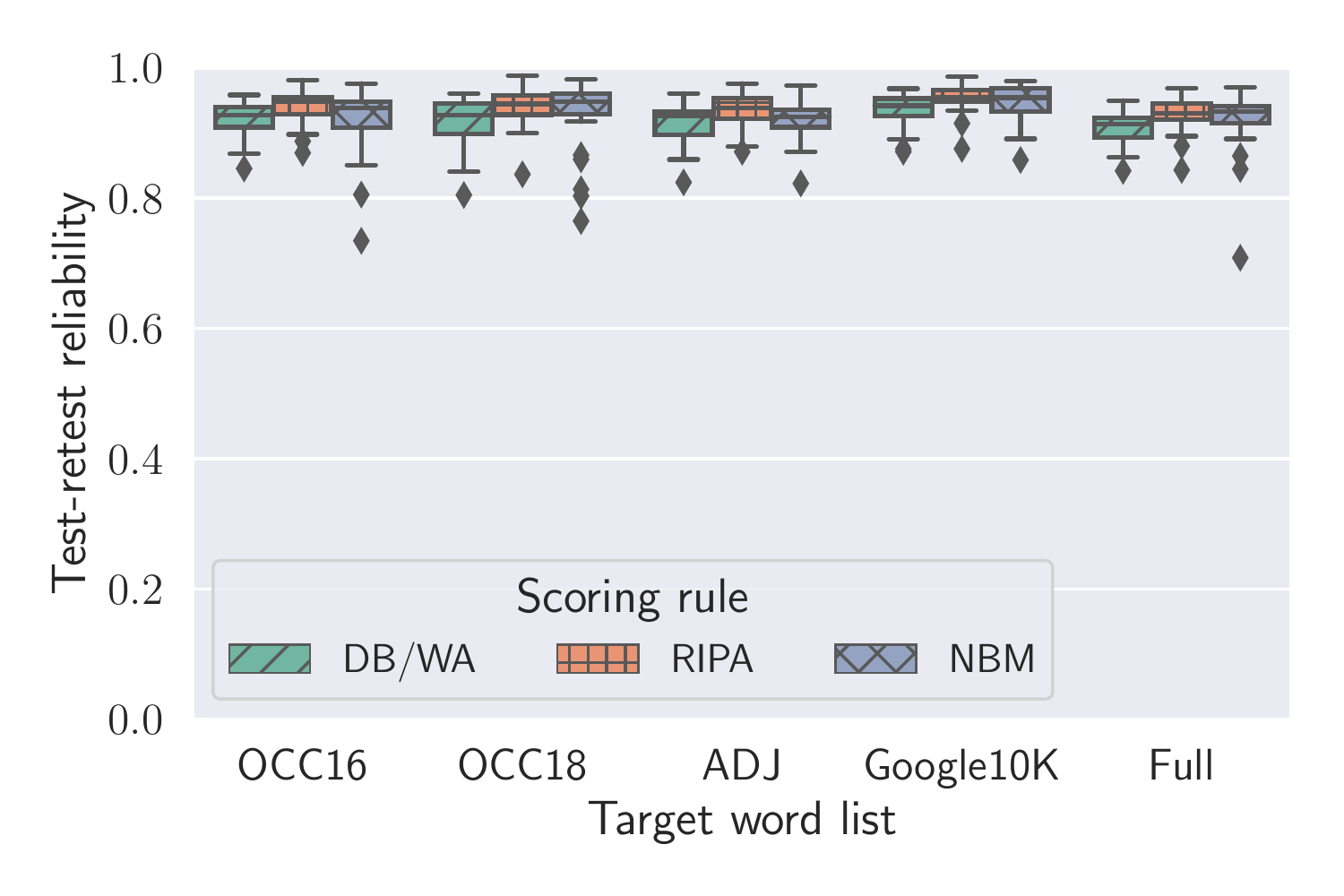}
    \caption{Test-retest reliability of gender base pairs. 
     The word embeddings are trained with SGNS on r/AskHistorians.}
\end{figure}

\begin{figure}[h!]
    \centering
    \includegraphics[width=0.48\textwidth]{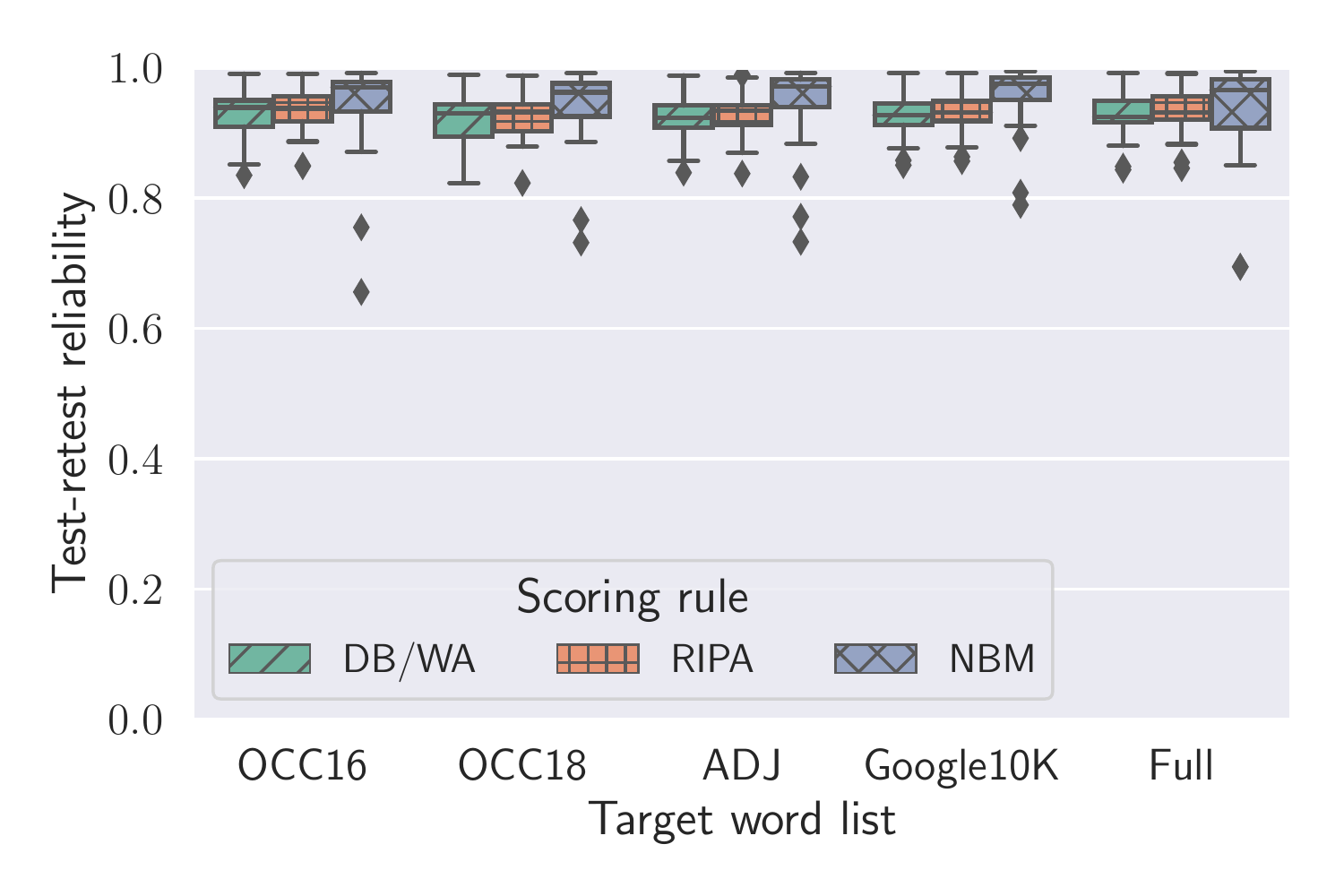}
    \caption{Test-retest reliability of gender base pairs. 
    The  word embeddings are trained with GloVe on r/AskHistorians.}
\end{figure}

\clearpage

\subsection{Inter-rater Consistency}\label{app:inter_rater_consistency_figure}

\begin{figure}[h!]
    \centering
    \includegraphics[width=0.48\textwidth]{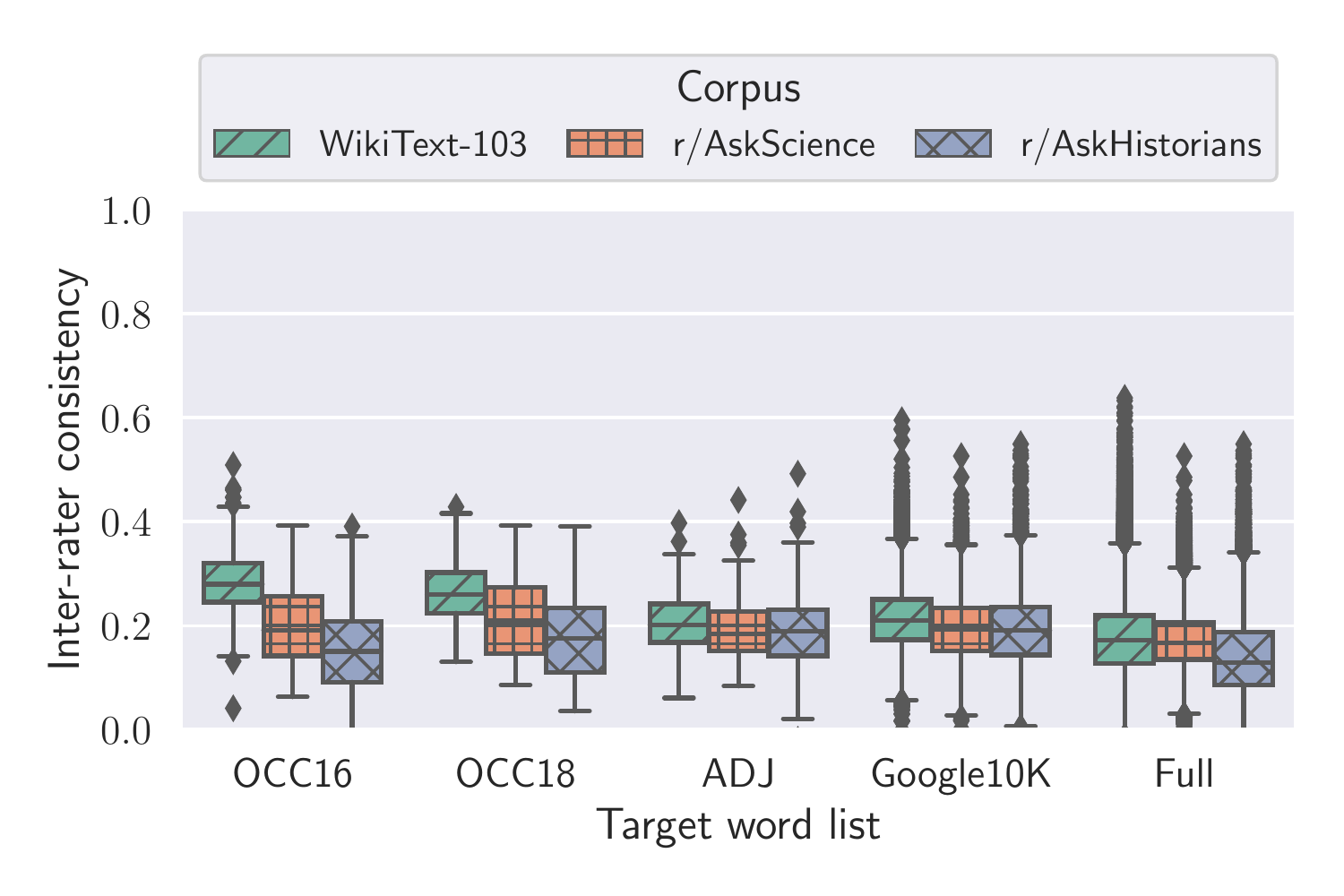}
    \caption{Inter-rater consistency of target words. 
     The word embeddings are trained with SGNS.}
\end{figure}

\begin{figure}[h!]
    \centering
    \includegraphics[width=0.48\textwidth]{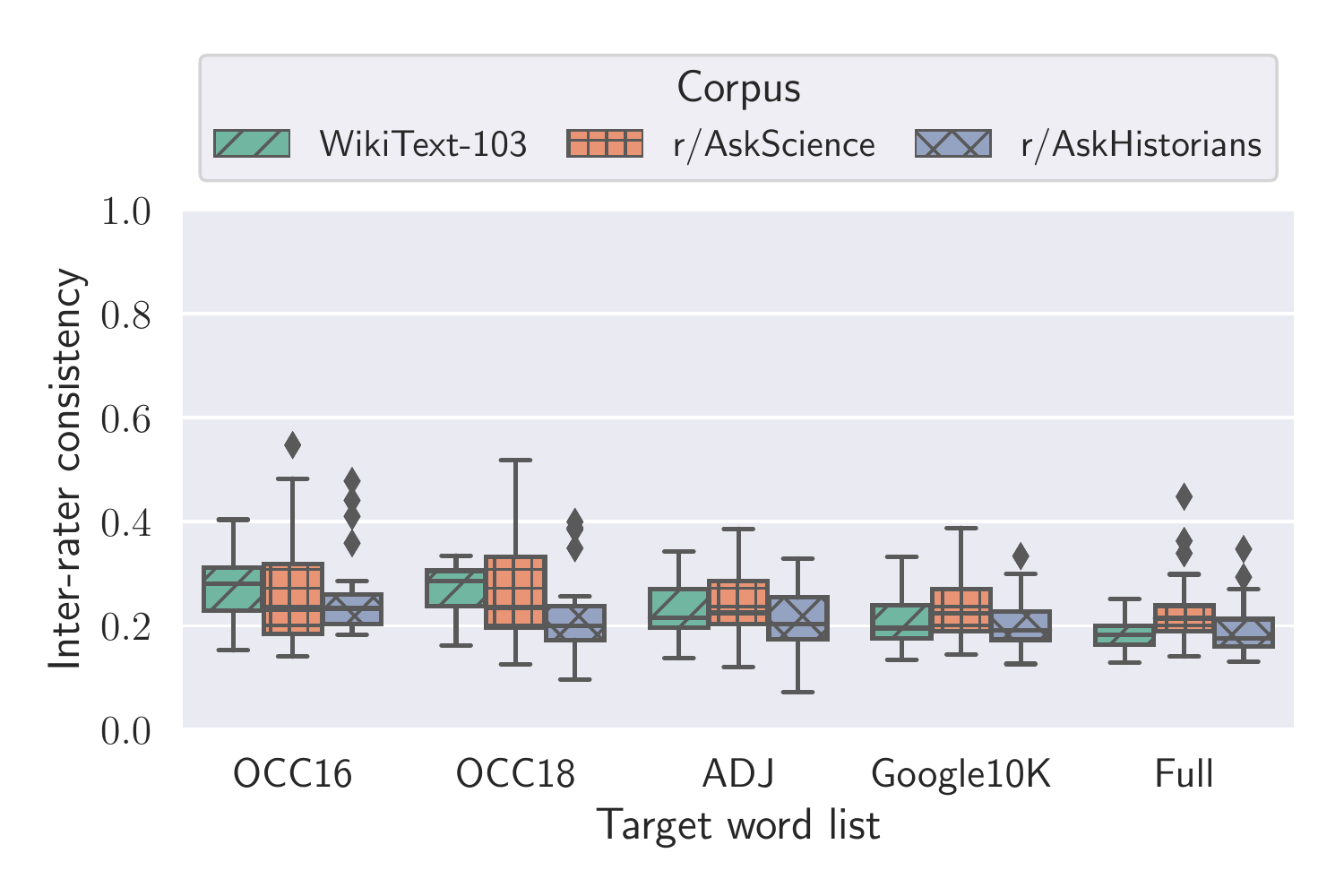}
    \caption{Inter-rater consistency of gender base pairs. 
    The  word embeddings are trained with SGNS.}
\end{figure}

\clearpage

\section{Effect of Hyper-parameters}\label{app:hyperparameter}

In our study, we did not fine-tune the hyper-parameters while training the word embeddings, 
because the main goal of this paper is not to compare different setups of word embedding algorithms.
In this section, we explore whether the results
are sensitive to the choice of hyper-parameters used for
training the word embeddings.
In the main paper, for SGNS, 
we use 300 dimensions and 5 iterations.
Here, we experiment with two different hyper-parameters on \emph{r/AskHistorians}, 
1) 3 iterations, and 2) 100 dimensions, respectively. 

\begin{figure}[h!]
    \centering
    \includegraphics[width=0.48\textwidth]{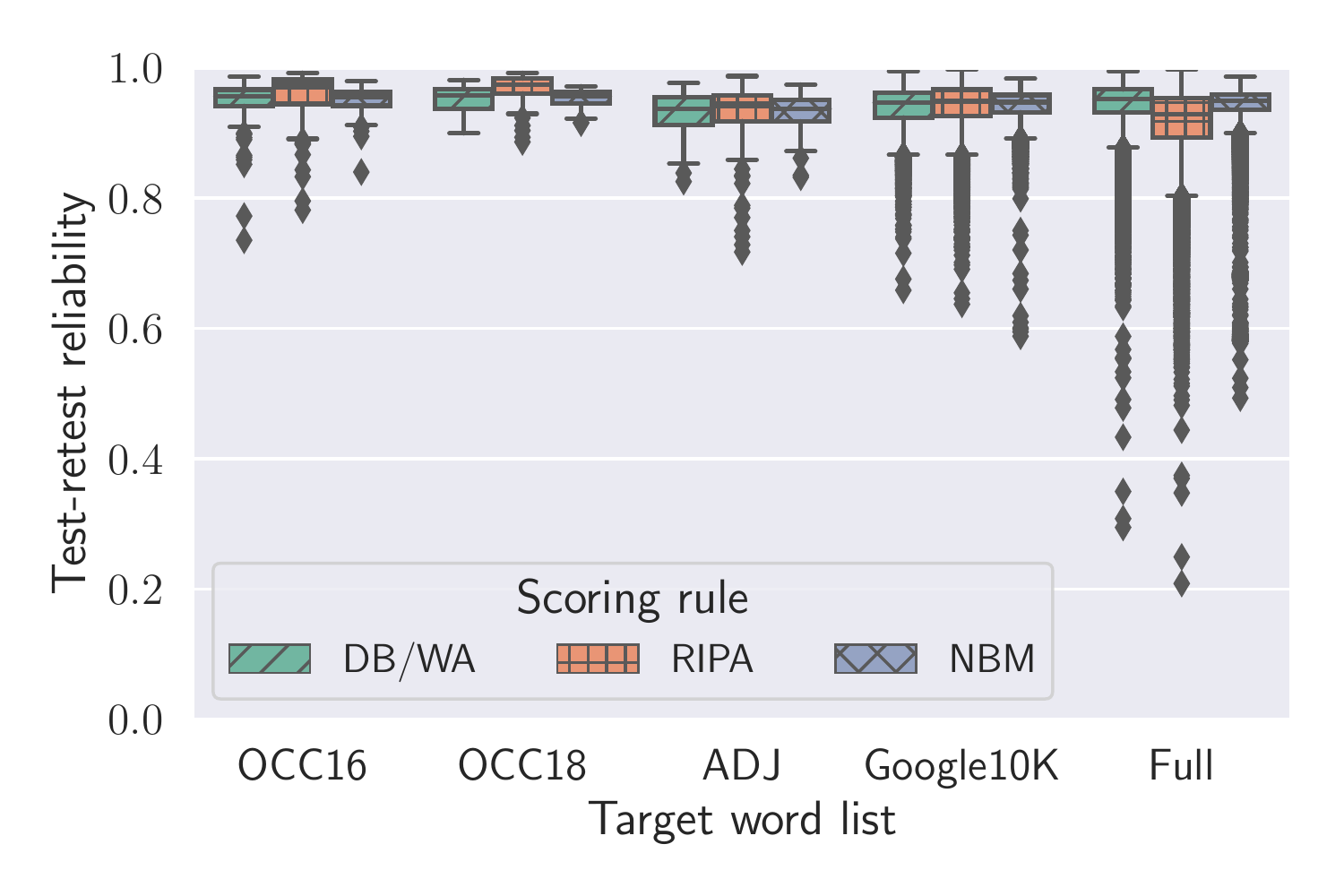}
    \caption{Test-retest reliability of target words of 
        100-dimensional SGNS word embeddings trained on \emph{r/AskHistorians}.}
    \label{fig:retest_100d_target}
\end{figure}

\begin{figure}[h!]
    \centering
    \includegraphics[width=0.48\textwidth]{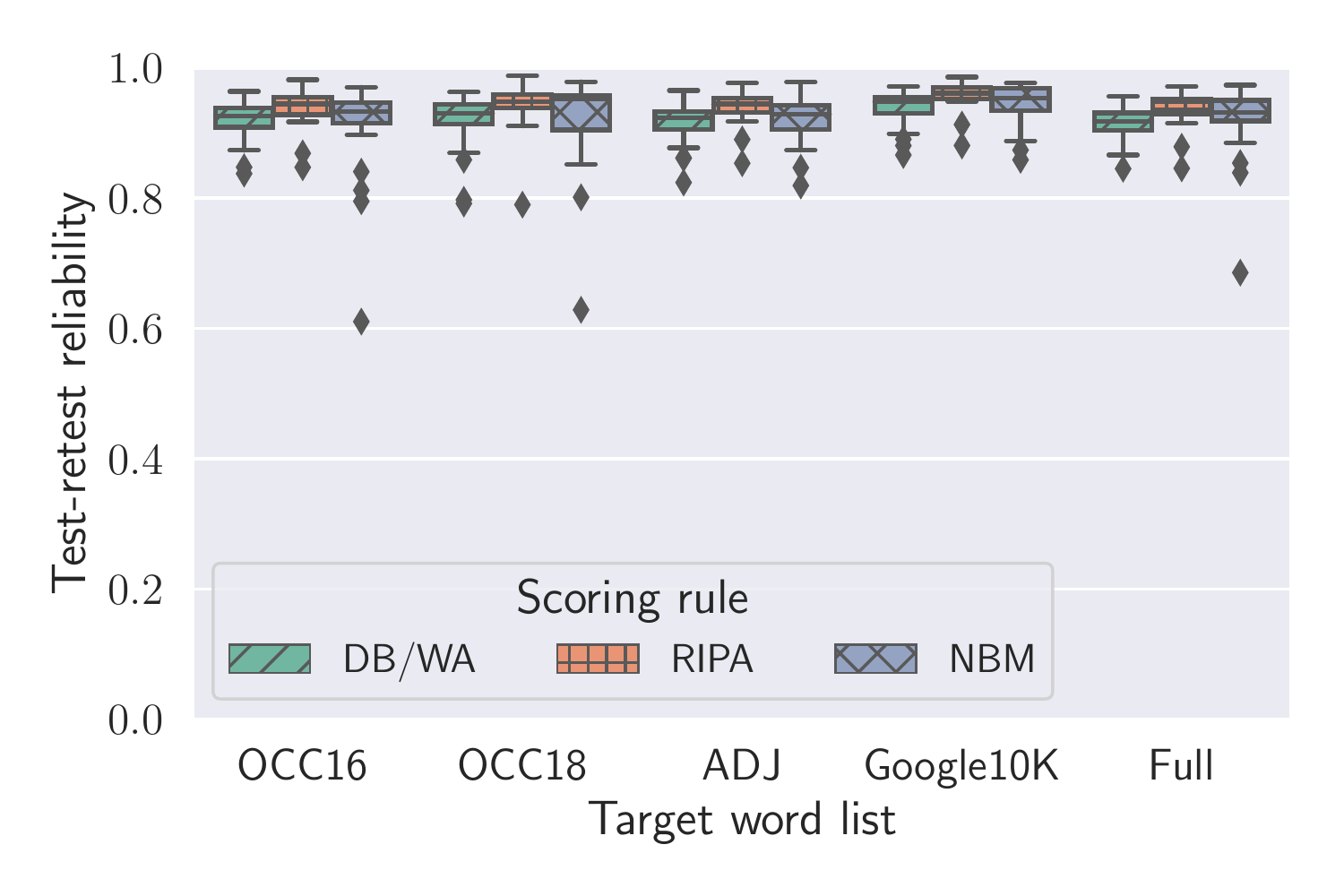}
    \caption{Test-retest reliability of gender base pairs of
        100-dimensional SGNS word embeddings trained on \emph{r/AskHistorians}.}
    \label{fig:retest_100d_gender}
\end{figure}

\begin{figure}[h!]
    \centering
    \includegraphics[width=0.48\textwidth]{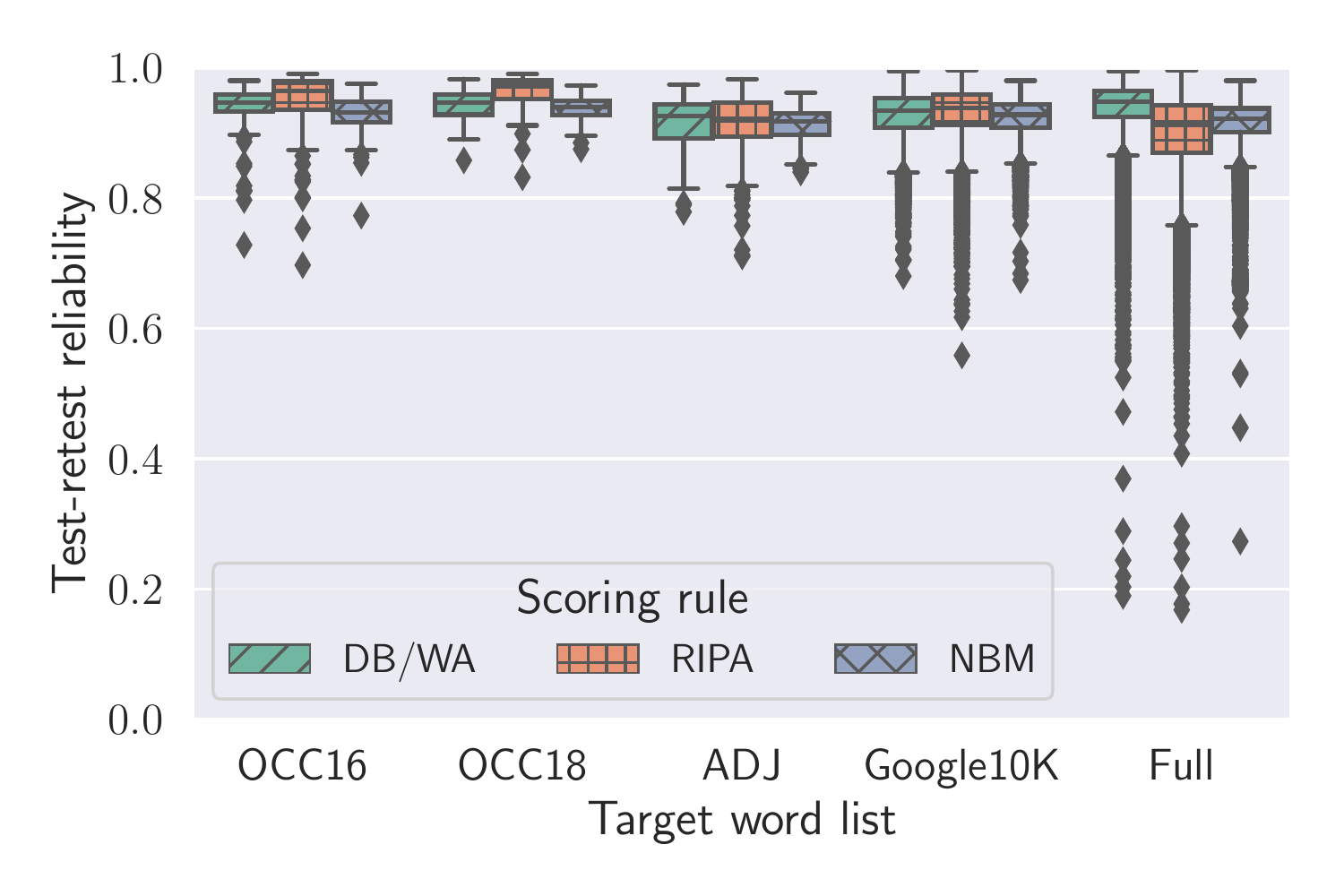}
    \caption{Test-retest reliability of target words of 
        SGNS word embeddings trained with three iterations on \emph{r/AskHistorians}.}
    \label{fig:retest_3e_target}
\end{figure}

\begin{figure}[h!]
    \centering
    \includegraphics[width=0.48\textwidth]{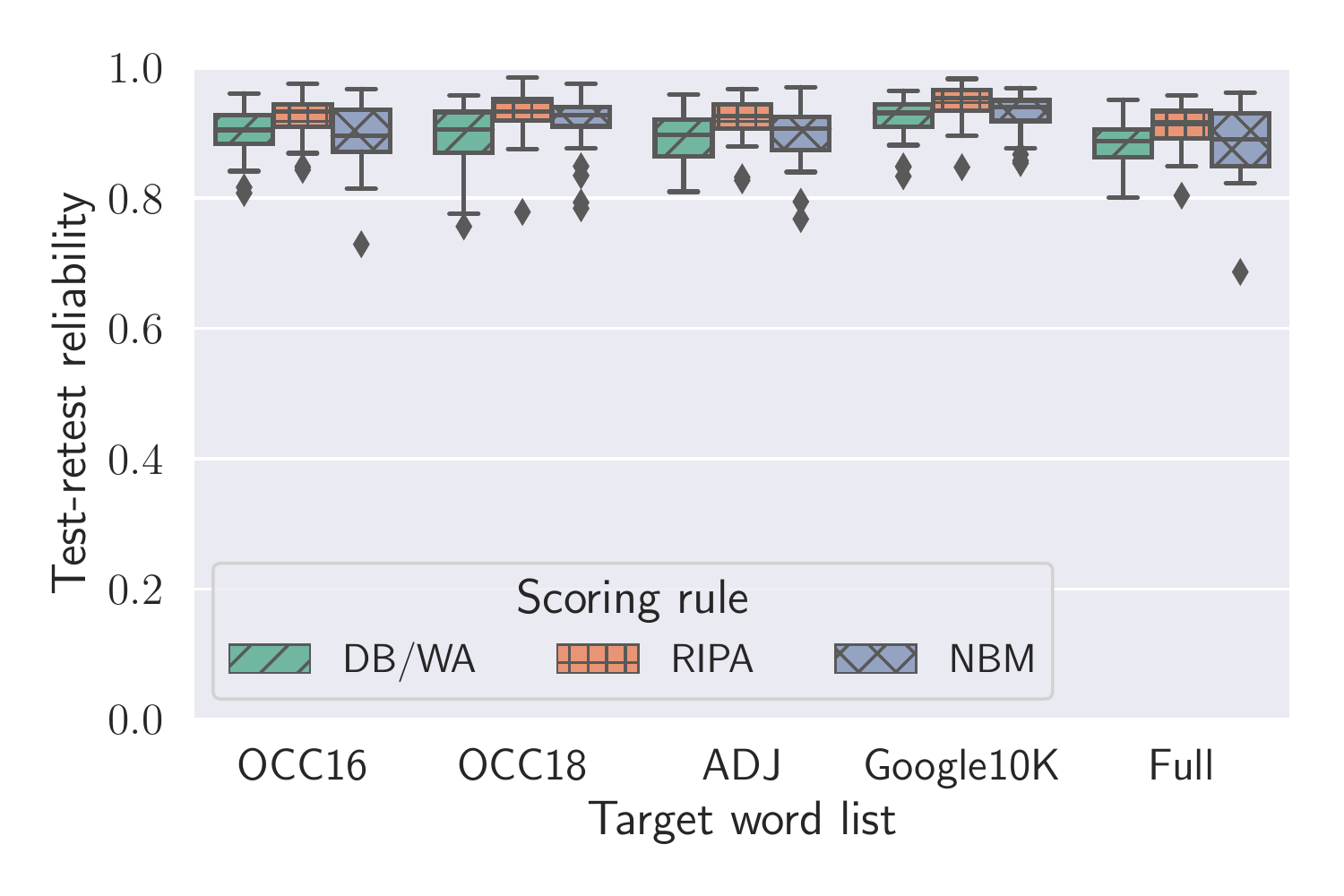}
    \caption{Test-retest reliability of gender base pairs of
        SGNS word embeddings trained with three iterations on \emph{r/AskHistorians}.}
    \label{fig:retest_3e_gender}
\end{figure}

\begin{figure}[h!]
    \centering
    \includegraphics[width=0.48\textwidth]{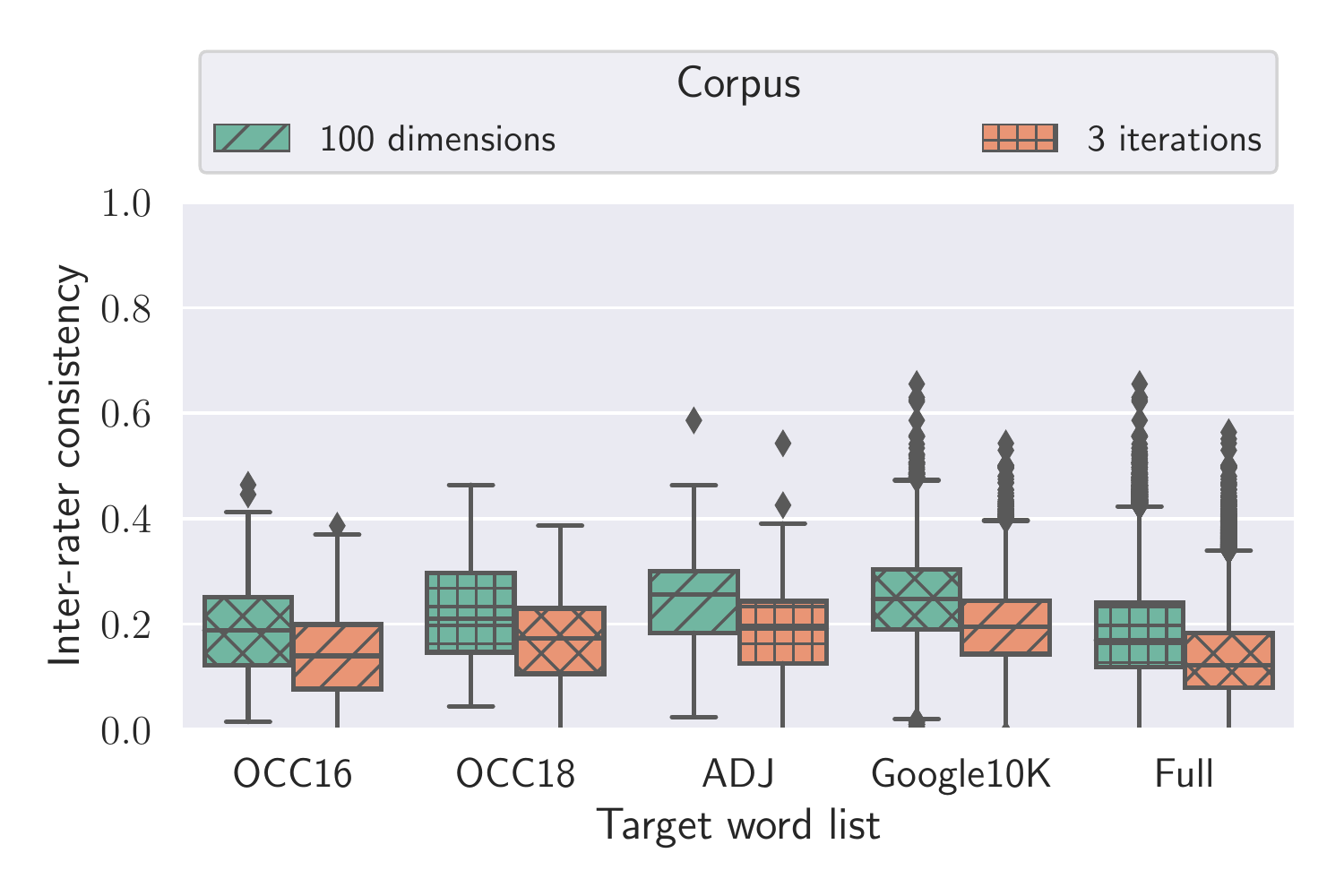}
    \caption{Inter-rater consistency of target words of 
        SGNS word embeddings trained with three iterations or 100 dimensions on \emph{r/AskHistorians}.}
    \label{fig:rater_3e_100d_target}
\end{figure}

\begin{figure}[h!]
    \centering
    \includegraphics[width=0.48\textwidth]{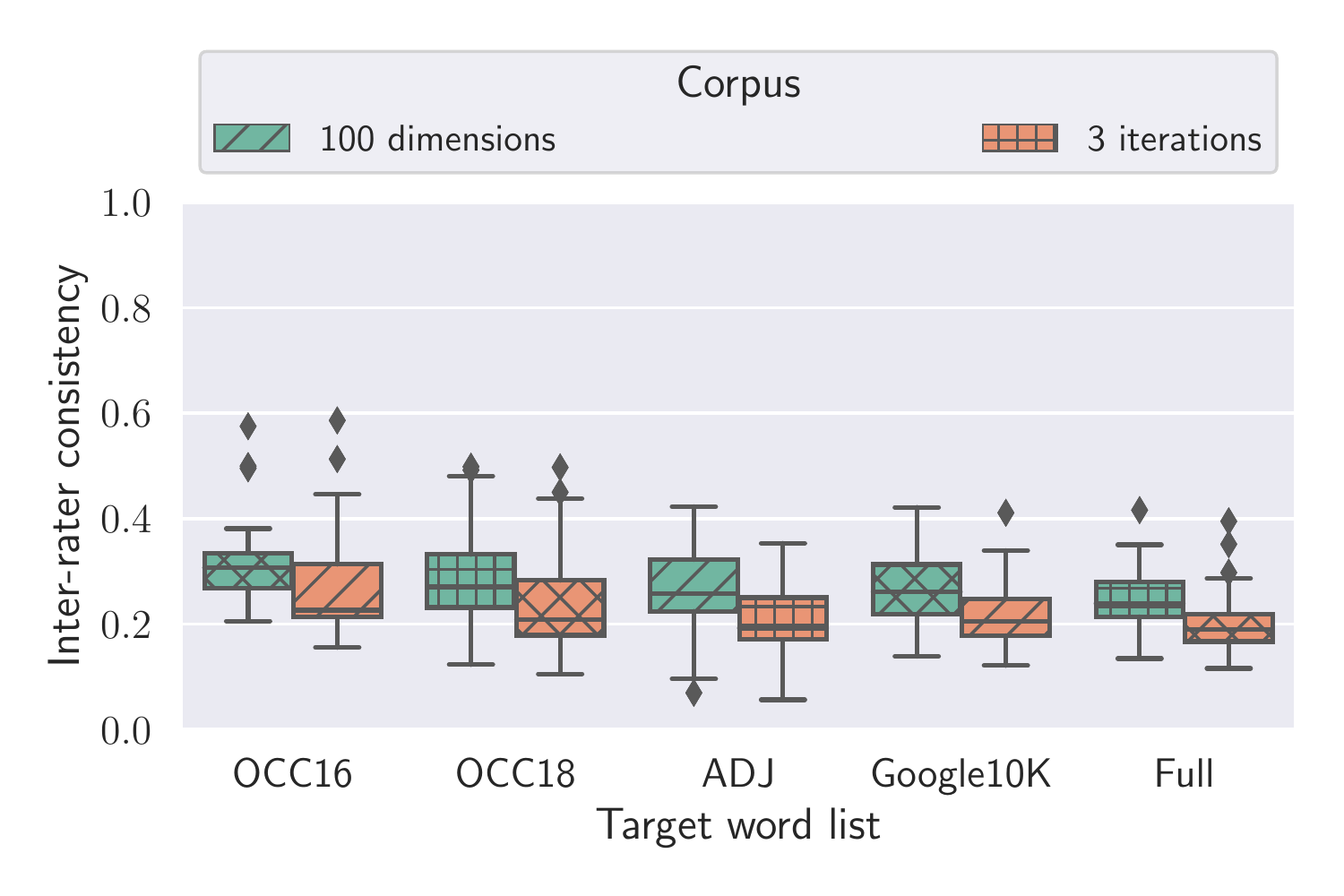}
    \caption{Inter-rater consistency of gender base pairs of 
        SGNS word embeddings trained with three iterations or 100 dimensions on \emph{r/AskHistorians}.}
    \label{fig:rater_3e_100d_gender}
\end{figure}

\begin{figure}[h!]
    \centering
    \includegraphics[width=0.48\textwidth]{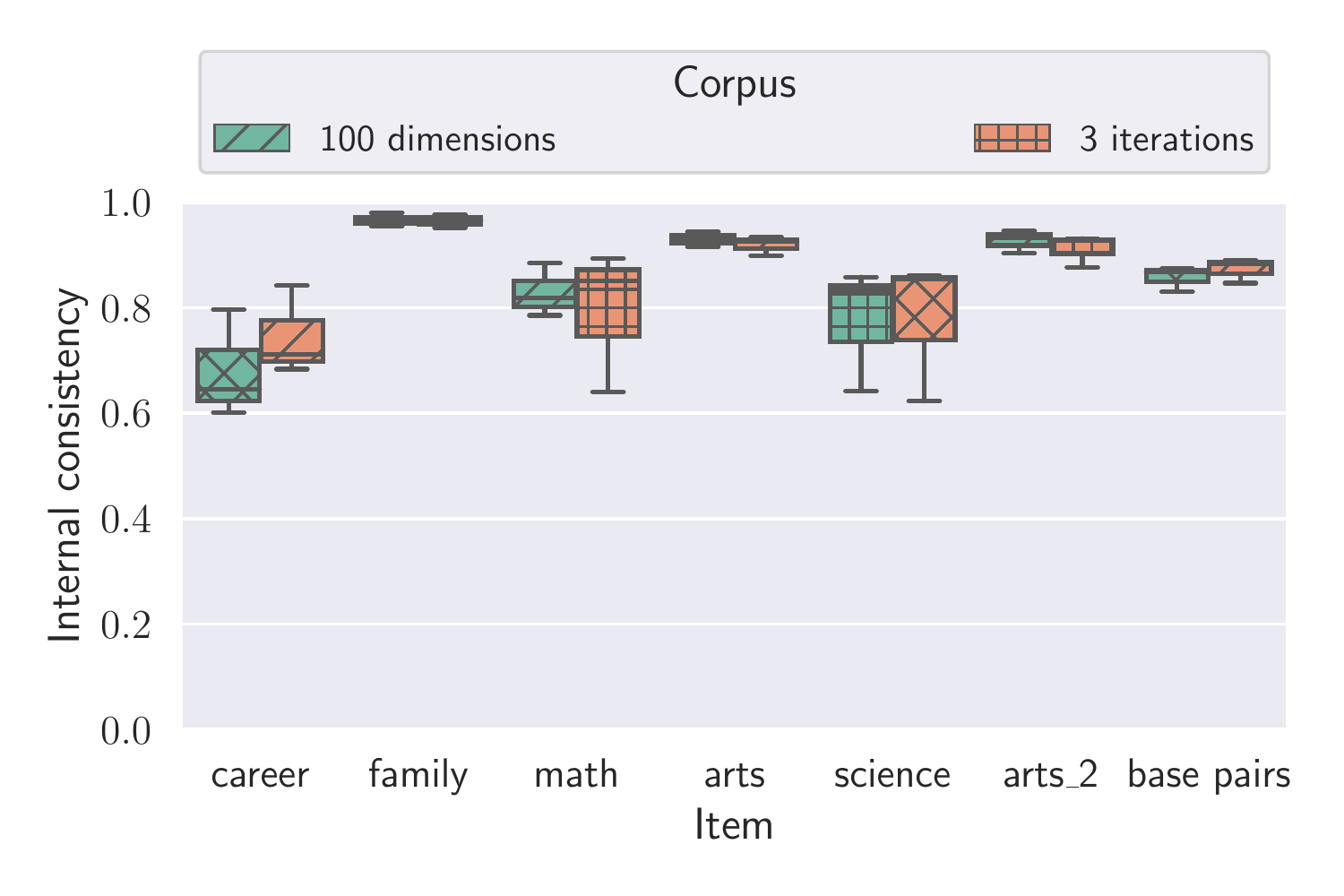}
    \caption{Internal consistency of 
        SGNS word embeddings trained with three iterations or 100 dimensions on \emph{r/AskHistorians}.}
    \label{fig:internal_3e_100d}
\end{figure}

The results are shown in Figures \ref{fig:retest_100d_target} to \ref{fig:internal_3e_100d}. 
Comparing with the default hyper-parameters, we observe that 
although the specific values of different types of reliability change, 
the overall trends remain the same. 

\clearpage
\section{Multilevel Regression}\label{app:mlr}
This section is a detailed description of the multilevel regression experiments in \S \ref{subsec:factors}.
\begin{table*}[t]
\centering
\begin{tabular}{c|cc|cc}
\toprule
            & \multicolumn{2}{c}{Test-retest} & \multicolumn{2}{c}{Inter-rater}\\ 
            & Estimate & $\Delta R^2$ & Estimate & $\Delta R^2$\\ \midrule
$\text{scoring rule}_\text{DB/WA}$ &  reference & 0.0040 & - & -\\
$\text{scoring rule}_\text{RIPA}$       & \textbf{-0.0102} &  & - & - \\
$\text{scoring rule}_\text{NBM}$  & -$0.0001$ &  & - & -\\
log freq         & \textbf{0.0062} & 0.0275 & \textbf{0.0241} & 0\\
log$^2$ freq    & \textbf{0.0023} & 0.0051 & \textbf{0.0008} & 0\\
log \#senses & \textbf{-0.0012} & 0.0002  & \textbf{-0.0007} & 0\\
$\text{PoS}_\text{adj}$  &  reference & 0.0093 & reference & 0.0022\\
$\text{PoS}_\text{adp}$        & \textbf{-0.0096} &  & \textbf{-0.0195} &\\
$\text{PoS}_\text{adv}$         & -0.0004 &  & \textbf{-0.0118}&\\
$\text{PoS}_\text{conj}$         & -0.0010 &  & \textbf{-0.0446}&\\
$\text{PoS}_\text{det}$         & \textbf{-0.0212} &  & \textbf{-0.0246}&\\
$\text{PoS}_\text{noun}$   & \textbf{0.0085} &  & \textbf{0.0057}&\\
$\text{PoS}_\text{num}$   & \textbf{-0.0025} &  & \textbf{-0.0886}&\\
$\text{PoS}_\text{pron}$         & -0.0039 &  & -0.0012&\\ 
$\text{PoS}_\text{prt}$ & 0.0037 &  & \textbf{-0.0125}&\\
$\text{PoS}_\text{verb}$ & 0.0001 &  & \textbf{-0.0056}&\\
$\text{PoS}_\text{x}$ & \textbf{-0.0031} &  & \textbf{-0.0205}&\\
NN Sim & \textbf{-0.0024} & 0.0120 &  \textbf{0.0011}& 0.0038\\
L2 norm & \textbf{-0.0118} & 0.0456 & \textbf{-0.0051}& 0\\
ES & \textbf{0.0457} & 0.1020 &  \textbf{0.0186}&0\\ \midrule
$R^2_{\text{fixed}}$& 0.3261 & - &0.0581&-\\
$R^2_{\text{corpus}}$& 0.0113 & - &0.0271&-\\
$R^2_{\text{algorithm}}$& 0.1802 & - &0.4307 &-\\
$R^2_{\text{total}}$ & 0.5177 & - &0.5160 &-\\ \bottomrule
\end{tabular}
\caption{Results of multilevel regression on 
    the test-retest and inter-rater reliability of target words.
    Estimates are standardized (bold if $p < 0.05$).
    $\Delta R^2$ is reduction in explained variance 
    when the corresponding factor is left out.
    $R^2_{\text{fixed}}$, $R^2_{\text{corpus}}$, $R^2_{\text{algorithm}}$ 
    and $R^2_{\text{total}}$ 
    refer to the explained variance of fixed factors 
    (i.e. word level features and scoring rules), 
    embedding training corpus used, 
    embedding training algorithm used, 
    total effects of all these three parts, 
    respectively. 
}
\label{tab:multilevel_full}
\end{table*}

\paragraph{Multilevel Models}

In a similar research setup, 
\citet{burdick2018factors} used variants of OLS regression 
to study influencing factors of word embedding stability. 
However, OLS regression has a strong assumption that 
the observations are unconditionally independent of one another. 
If this assumption is violated, 
standard errors of regression coefficients will be underestimated, 
leading to an overstatement of statistical significance. 

In our study, 
our observations (reliability scores of target words) 
are nested within different corpora and word embedding algorithms. 
As a result, 
the residual variance is naturally partitioned 
into a within-corpus-algorithm component 
and a between-corpus-algorithm component. 
If the between-component is not explicitly accounted for, 
the observations will be correlated 
(thus not independent anymore) within a corpus or algorithm. 

We therefore use \textbf{multilevel} regression 
instead of OLS regression (or its variants). 
Multilevel regression 
(or mixed-effect models, hierarchical models, among many other names) 
accounts for the grouping/hierarchical structure of data 
by explicitly modelling residuals at different levels of the data. 
In this way, the model not only relaxes 
the assumption of unconditional independent data, 
but also estimates group effects. 
The latter is especially beneficial 
because it allows us to quantify 
how much variance in the data is explained by 
the corpora and the algorithms, 
in addition to word-level features. 

Formally, our multilevel regression model is:

$$
\begin{aligned}
  y_{i(a,c)}  &=X_{i(a,c)}\beta+\nu_{0a}+\mu_{0c}+\epsilon_{i(a,c)} \\
    \nu_{0a}  &\sim N \left(0, \sigma^2_{\nu_0} \right) \\
    \mu_{0a}  &\sim N \left(0, \sigma^2_{\mu_0} \right) \\
    \epsilon_{0a}  &\sim N \left(0, \sigma^2_{\epsilon_0} \right)
\end{aligned}
$$

where $y_{i(a,c)}$ refers to each reliability score of a target word $i$ nested within an algorithm $a$ and corpus $c$. 
$X_{i(a,c)}$ is a row vector containing observations on the word-level explanatory factors with a leading element of one.
$\beta$ is a column vector of parameter estimates of those factors.
$\nu_{0a}$, $\mu_{0a}$, $\epsilon_{0a}$ are the residual error terms for algorithm $a$, corpus $c$ and target word $i$. 
They are assumed to be normally distributed around zero with their own variances.

Note that we can not conduct such regression analyses for reliability scores on the level of queries, gender base pairs or scoring rules due to limited sample sizes (i.e. only 6 queries, 23 gender base pairs and 3 scoring rules).

\paragraph{Feature Selection}

We collect a wide range of features for the regression model. 
They can be divided into two categories: 
word-level features and group-level features. 

We use 
several word-level features. 
First, we consider the natural logarithm of the number of synsets in WordNet (log \#senses).  
We regard it as an approximation of 
the number of senses of words \citep{burdick2018factors}. 
Then, we use the most common PoS tags of words (PoS). 
We use it to represent the syntactic roles of words in sentences 
\citep{burdick2018factors,pierrejean-tanguy-2018-predicting}. 
We call these two features word-intrinsic features 
because they are unrelated to the training corpora or algorithms. 

We also include the natural logarithm of the frequencies of words (log freq). 
Frequency is found to influence the stability of word embeddings 
by \citet{DBLP:conf/coling/HellrichH16}. 
Then, we consider its squared value as well (log$^2$ freq),  
because the relationship between word frequency and 
its corresponding embedding stability appears to be quadratic in our data. 
We refer to both words' frequencies and their squares 
as corpus-related features 
since they are only related to the training corpora. 

Then, we use several embedding-related features. 
First, we consider a word's cosine similarity with its nearest neighbour 
(NN Sim, \citealt{pierrejean-tanguy-2018-predicting}). 
Intuitively, 
stable embeddings should have stable nearest neighbours. 
We also use the L2 norm of word embeddings 
(L2 Norm, \citealt{pierrejean-tanguy-2018-predicting}). 
Third, we use embedding stability (ES) as a predictor too. 
Previous studies 
\citep{burdick2018factors,DBLP:conf/coling/HellrichH16,antoniak2018evaluating} 
usually measure the stability of word embeddings 
by the changes of their nearest neighbours. 
However, this method is insensitive to minor changes 
in the embedding space. 
Instead, 
we adopt a different method that detects the changes 
of lexical semantics 
\citep{hamilton-etal-2016-diachronic,tan-etal-2015-lexical,Kulkarni:langchange},
to measure embedding stability in this paper. 
Specifically, 
given two word embedding models $W_1, W_2$, 
we fit a transformation matrix $Q$ , where 
\begin{IEEEeqnarray}{rCl}
    Q = \argmin_{Q^{\top}Q=I} \|W_1 - QW_2\|_{F}.\notag
\end{IEEEeqnarray}
Then the stability of word $w$ is given by 
the cosine similarity of $(w_1, Qw_2)$, 
where $w_1$ and $w_2$ are the 
corresponding word vectors of $w$ in 
word embeddings $W_1$ and $W_2$. 

Furthermore, for the regression model on test-retest reliability, we also include scoring rules as a fixed factor. 
Scoring rules can also be considered as a grouping factor, as the bias scores are also clustered within scoring rules.
However, this approach would not yield specific effect estimates for each of the three scoring rules, 
but instead only a variance estimate for all the scoring rules as a whole.
Because we are interested in comparing the effects of the three scoring rules on target words' test-retest reliability scores, 
we explicitly model scoring rules as a fixed factor.

Lastly, we include word embedding training corpora 
and word embedding training algorithms as the two group-level factors. 

\paragraph{Results}
The full results are in Table \ref{tab:multilevel_full}. 
The interpretation of the table is the same as in $\S \ref{subsec:factors}$. 
The main difference between this table and Table \ref{tab:multilevel} is that the parameter estimates of the PoS feature are no longer omitted. 
We can see that this feature as a whole explains less than 1\% of the variation in both the test-retest reliability and the inter-rater consistency of target words.
This suggests that in the presence of other features, 
PoS is not a very important factor.
Nevertheless, we do see some statistically significant and moderately sized parameter estimates of PoS categories. 
For instance, compared to adjectives,
determiners tend to score on average 0.0212 lower on test-retest reliability. 
Numerals also score on average -0.0886 lower on inter-rater consistency than do adjectives.

\end{document}